%% file: main.tex
\def\year{2022}\relax
\documentclass[letterpaper]{article} 
\usepackage{aaai22}  
\usepackage{times}  
\usepackage{helvet}  
\usepackage{courier}  
\usepackage[hyphens]{url}  
\usepackage{graphicx} 
\def\UrlFont{\rm}  
\usepackage{natbib}  
\usepackage{caption} 
\DeclareCaptionStyle{ruled}{labelfont=normalfont,labelsep=colon,strut=off} 
\usepackage{hyperref}       
\usepackage{url}            
\usepackage{booktabs}       
\usepackage{amsfonts}       
\usepackage{nicefrac}       
\usepackage{microtype}      
\usepackage{xcolor}         

\usepackage[export]{adjustbox}
\usepackage{adjustbox}
\usepackage{pgfplots}
\usepackage{subcaption}
\usepackage{tikz}
\usetikzlibrary{arrows,decorations.pathmorphing,backgrounds,positioning,fit,petri, decorations.pathreplacing,shadows,calc}
\usepackage{algorithm}
\usepackage{algorithmic}

\usepackage{selectp}
\newlength{\mylength}
\makeatletter
\newcommand{\mycfs}[1]{%
  \normalsize
  \@defaultunits\mylength=#1pt\relax\@nnil
  \edef\@tempa{{\strip@pt\mylength}}%
  \ifx\protect\@typeset@protect
     \edef\@currsize{\noexpand\mycfs\@tempa}
  \fi
  \mylength=1.2\mylength
  \edef\@tempa{\@tempa{\strip@pt\mylength}}%
  \expandafter\fontsize\@tempa
  \selectfont
}
\makeatother

\usepackage{tabularx}
\newcolumntype{P}[1]{>{\centering\arraybackslash}p{#1}}
\newcolumntype{M}[1]{>{\centering\arraybackslash}m{#1}}

\usepackage{mathtools} 
\usepackage{moresize}
\usepackage{amsmath} 
\usepackage{amsthm}
\usepackage{amssymb}
\usepackage{times}
\usepackage{graphicx}
\usepackage{latexsym}

\def\SPSB#1#2{\rlap{\textsuperscript{{#1}}}\SB{#2}}

\def\SB#1{\textsubscript{{#1}}}

\usepackage[switch]{lineno}
\usepackage{color}

\newtheorem{definition}{Definition}

\newtheorem*{rem*}{Remark}
\newtheorem{prop}{Proposition}

\newtheorem{example}{Example}

\usepackage{newfloat}
\usepackage{listings}
\floatstyle{ruled}
\newfloat{listing}{tb}{lst}{}
\floatname{listing}{Listing}

\title{Systematic Generalisation of Temporal Tasks\\ through  Deep Reinforcement Learning}

\author {
    Borja G. Le\'on,\textsuperscript{\rm 1}
    Murray Shanahan, \textsuperscript{\rm 1}
    Francesco Belardinelli \textsuperscript{\rm 1}
}
\affiliations {
    \textsuperscript{\rm 1} Department of Computer Science, Imperial College London\\
    
{b.gonzalez-leon19, m.shanahan, francesco.belardinelli}@imperial.ac.uk
}

\begin{document}
\maketitle
\begin{abstract}
  This work introduces a neuro-symbolic agent that combines deep reinforcement learning (DRL) with temporal logic (TL) to achieve systematic zero-shot, i.e., never-seen-before, generalisation of formally specified instructions. In particular, we present a neuro-symbolic framework where a symbolic module transforms TL specifications into a form that helps the training of a DRL agent targeting generalisation, while a neural module learns systematically to solve the given tasks. We study the emergence of systematic learning in different settings and find that the architecture of the convolutional layers is key when generalising to new instructions. We also provide evidence that systematic learning can emerge with abstract operators such as negation when learning from a few training examples, which previous research have struggled with.
  

\end{abstract}

\section{Introduction} \label{sec:intro}
\input{introduction.tex}

\section{Systematic generalisation of formal tasks} \label{sec:backg}
\input{Background} \label{sec:Background}
\section{Solving zero-shot instructions in TTL} \label{sec:TTL}
\input{TTL}

\section{Experiments} \label{sec:experiments}
\input{experiments}

\section{Related work}  \label{sec:relatedWork}
\input{relatedWork}

\section{Discussion and conclusions} \label{sec:discussion}
\input{Conclusions}

\bibliography{utils/Bib}

\clearpage
\appendix
\input{appendix}

\end{document}

%% file: introduction.tex
Systematic generalisation (also called combinatorial 
generalisation \cite{lake2019compositional}) concerns the human ability of compositional 
learning, that is, 
the algebraic capacity of understanding and 
executing novel utterances by combining already known primitives \cite{fodor1988connectionism, chomsky2002syntactic}. For instance, once a human agent understands the instruction ``get {\em wood}" and the meaning of {\em iron}, they will understand the task ``get {\em iron}".  This is a desirable feature for a computational model, as it suggests that once the model is able to understand the components of a task, it should be able to satisfy tasks with the same components in possibly different combinations.

The ability of neural-based agents to learn systematically and generalise beyond the training environment is a recurrent point of debate in machine learning (ML). In the context of autonomous agents solving human-given instructions, deep reinforcement learning (DRL) holds considerable promise \cite{co2019meta, luketina2019survey}. Previous studies have shown that DRL agents can exhibit some degree of systematic generalisation of natural language \cite{yu2018interactive, bahdanau2018systematic, hill2021grounded}. In parallel, frameworks based on structured or formal languages to instruct DRL agents are also gaining momentum \cite{yuan2019modular, de2019foundations, icarte2019learning}. Formal languages offer several desirable properties for RL including unambiguous semantics, and compact compositional syntax, which is particularly relevant when targeting safety-aware applications \cite{alshiekh2018safe, jothimurugan2019composable, simao2021alwayssafe}. However, works with formal languages have traditionally focused on solving training instructions \cite{icarte2018using, hasanbeig2021deepsynth, hammond2021multi} or rely on new policy networks when tackling novel instructions \cite{kuo2020encoding}, with its associated computational burden.

In order to progress towards agents that execute zero-shot, i.e., unseen, formal instructions it is desirable to design frameworks that facilitate the emergence of systematic learning in DRL. There exists substantial evidence that DRL agents can learn systematically only when the right drivers are present \cite{lake2019compositional, mao2019neuro}. Unfortunately, the focus on learning from natural instructions has hindered the study of drivers for systematicity with logic operators, e.g., negation or disjunction. Only \cite{hill2020environmental} provides evidence of DRL agents with some degree of systematicity facing negated zero-shot instructions. This work suggests that systematic learning can emerge from as few as six instructions when handling positive tasks, but that a much larger number of instructions ($\sim 100$) is a major requirement when aiming abstract operators such as negation. This constraint would be a strong burden for DRL agents, meaning they could only generalise abstract concepts when rich and diverse environments are available.


\paragraph{Contribution} In this work we aim to answer the question: {\em can systematic learning emerge in DRL agents with abstract operators when learning from a few ($\sim 6$) examples?} 
Specifically, we make the following contributions:
\begin{itemize}
    \item We introduce a novel neuro-symbolic (NS) framework aiming for agents that exhibit systematic generalisation. In this framework, a formal language aids neural-based agents that target 
    zero-shot instructions, while the symbolic module facilitates generalisation to longer than training instructions.
    \item We provide evidence of emergent systematic generalisation with abstract operators -- including negation -- from six training instructions.
    \item We find that the architecture of the convolutional layers (c.l.) -- a background feature in previous literature -- can be key to generalise with zero-shot tasks.
\end{itemize}

%% file: Background.tex
We train agents that execute instructions expressed in temporal logic (TL) and operate in partially observable (p.o.) environments. Our target is to assess whether systematic learning is emerging in such environments while stressing that the CNN architecture is key facilitating generalisation when the variety on training examples is limited. This section briefly introduces the concepts needed to present our work.

\paragraph{Reinforcement Learning.} In RL p.o.~environments are typically modelled as a POMDP. 
\begin{definition}[POMDP]
A {\em p.o.~ Markovian Reward Decision Process} is a
tuple $\mathcal{M}=\langle S, A, P, R, Z, O, \gamma\rangle$
where
(i) $S$ is the set of {\it states} $s, s', \ldots$.
(ii) $A$ is the set of {\em actions} $a, a', \ldots$.
(iii) $P\left(s^{\prime} | s, a\right): S \times
    A \times S \rightarrow[0,1]$ is the (probabilistic) {\em transition function}.
(iv) $R(s, a, s): S \times A \times S \rightarrow \mathbb{R}$ is the {\em Markovian reward function}.
 (v)   $Z$  is the set of {\em observations} $z, z', \ldots$.
  (vi)  $O(s, a, s): S \times A \times S \rightarrow Z$ is the {\em observation function}.
(vii)    $\gamma \in[0,1)$ is the {\em discount factor}.
\end{definition}

\looseness=-1
At every timestep $t$, the agent chooses an action $a_t \in A$ updating the current state from $s_t$ to $ s_{t+1} \in S$  according to $P$.
Then, $R_t= R(s_t, a, s_{t+1})$ provides the reward associated with the transition, and $O(s_t,a_t,s_{t+1})$ generates an observation $o_{t+1}$ that will be provided to the agent. Intuitively, the goal of the learning agent is to choose the policy $\pi$ that maximize the expected discounted reward from any state $s$. Given the action-value function $Q^{\pi}(s, a)=\mathbb{E}\left[R_{t} | s_{t}=s, a\right]$, RL searches for an optimal policy whose actions are determined by the optimal action-value function $Q^{*}(s, a) = \max_{\pi} Q^{\pi}(s, a)$. For more detail we refer to \citet{sutton2018reinforcement}.

\paragraph{Solving tasks in temporal logic.} Temporal Logic is commonly used to express (multi-task) temporally extended goals \cite{huth2004logic}. Intuitively, this means that goals in TL are naturally modelled as non-Markovian, as the evaluation of a formula in TL depends on the whole history of states and actions, which we will refer to as {\em traces} $\lambda, \lambda', \ldots$. Then, a p.o.~problem where RL is used to solve TL goals (an RL-TL problem) is modelled as a {\em p.o.~Non-Markovian Reward Decision Process} (PONMRDP), which is a tuple $\mathcal{M}=\langle S,  A,
 P, \overline{R}, \phi, Z, O, \gamma\rangle$ where $S, A, P, Z, O \text{ and } \gamma$ are defined as in a POMDP, $\phi$ is the instruction expressed as a TL formula and $\overline{R}: (S \times A \times S)^{*} \rightarrow \mathbb{R}$ is the {\em non-Markovian reward function} associated with the satisfaction of $\phi$, which in our context is a TTL formula (see Sec.~\ref{sec:TTL}). Since RL is not designed to work with non-Markovian rewards \cite{sutton2018reinforcement}, RL-TL frameworks extend the state space $S$ with a representation of the current task to be solved, so that the PONMRDP is transformed into an equivalent extended POMDP (see \citet{bacchus1996rewarding, toro2018teaching} for references).

\paragraph{Convolutional neural network (CNN).} A CNN is a specialized type of neural network model \cite{lecun2015deep} designed for working with visual data. Particularly, a CNN is a network with one or more convolutional layers (c.l.). Relevant features of c.l. for this study include {\em kernels}, a matrix of weights which is slid across an input image and multiplied so that the output is enhanced in a desirable manner, and {\em stride}, which the measures the step-size of the kernel slid.
After the last convolutional layer, the output is a tensor which we refer to as {\em visual embedding}. We point to \citet{goodfellow2016deep} for a complete description of CNNs.

\looseness=-1
\paragraph{Evaluating the emergence of systematic generalisation.} 
As anticipated in Sec.~\ref{sec:intro}, we aim to assess whether a neuro-symbolic DRL agent is able to generalize systematically from TL instructions. Consequently, \emph{we are not interested in the raw reward gathered by the agents}, but rather on whether the agent is correctly following new instructions. To evaluate this, we work in two procedural gridworld settings where agents need to fulfill reachability goals, in line with previous literature \cite{hill2020environmental, icarte2019learning, kuo2020encoding}. In our experiments agents are trained with instructions such as ``get gold" (reach gold) or ``get something different from mud" (i.e., reach an object that is not mud). Then, we evaluate the agents with unseen instructions, e.g., ``get something different from iron then get iron" by navigating in the given order to the positions where ``an object different from iron" and ``and iron" are. It is evident that having a learning agent achieving significantly better rewards than a random walker does not imply systematic learning, e.g., the agent could accumulate larger rewards just by navigating better than a random walker and reaching all the objects promptly. For this reason, after training, we evaluate generalisation in settings called binary choice maps (BCMs). In these test maps, agents are evaluated with reliable instructions, i.e., when we give zero-shot instructions that point to an object granting a positive reward, or deceptive instructions, i.e., zero-shot instructions that point to objects granting a penalisation. Since during training we always provide reliable instructions, agents that show systematic learning -not complete systematicity but at least similar to closely related works \cite{bahdanau2018systematic, hill2020environmental}- should accumulate significantly higher rewards with zero-shot reliable instructions than with deceptive ones (around two times higher or greater, detailed in Sec.~\ref{sec:experiments}).

%% file: TTL.tex
\looseness=-1
We present a framework to tackle complex zero-shot temporal specifications by relying on 3 components detailed in this section: 1) a formal language, whose atoms are designed to facilitate the systematic learning of a neural-based module; 2) a symbolic module, whose role is to 
decompose specifications into sequences of simpler tasks, and to guide a neural module on the current task to execute; 3) a neural module consisting of a DRL agent designed to learn systematically from the tasks fed by the symbolic module.

\subsection{Task temporal logic} \label{subsec:TTL}

We start by defining a novel TL language
whose atomic components are used to train a DRL algorithm. 
Task temporal logic (TTL) is an expressive, learning-oriented TL
 language interpreted over finite traces, i.e, over finite episodes. Given a set $AT$ of atomic tasks $\alpha, \beta, \ldots$, the syntax of TTL is defined as follows: 
\begin{definition}[TTL] 
\looseness=-1
Every formula $\phi$ in TTL is built from atomic tasks $\alpha \in AT$ by using negation ``$\sim$" (on atoms only), sequential composition ``$;$" and non-deterministic choice of atoms ``$\lor$" and formulae ``$\cup$":
\begin{eqnarray*}
T & :: = &   \alpha \mid \alpha \sim \;\, \mid \alpha \lor \alpha \\
\phi \enskip & ::= & \enskip  T \mid  \phi ; \phi \mid \phi \cup \phi
\end{eqnarray*}
\end{definition} 

Intuitively, an atomic task $\alpha$ corresponds to a reachability goal in the TL literature, in the sense that the fulfilment condition associated with $\alpha$ will eventually hold. Task $\alpha \sim$ encapsulates a special form of negation; informally it means that something different from $\alpha$ will eventually hold. Choosing this form of negation allows us to contrast our results with those of previous studies of this operator in natural language \cite{hill2020environmental}. Note that the negation symbol is intentionally positioned after the atomic task. We found that this feature helps during training with visual instructions, as it forces the neural module (Sec.~\ref{subsec:NeuralMod}) to process the same number of symbols to distinguish negative from positive tasks (we refer to Sec.~\ref{subsec:Appendixnegation} in the supplemental material for further detail). Formulae $\phi; \phi'$ intuitively express that $\phi'$ follows $\phi$ in sequential order; whereas $\phi \cup \phi'$ means that either $\phi$ or $\phi'$ holds. We then also introduce an abbreviation for the concurrent (non-sequential) composition $\cap$ as follows: $\phi \cap \phi' \equiv (\phi; \phi') \cup (\phi';\phi)$. We say that our language TTL is {\em learning-oriented} as its logical operators and their positions in formulae are so chosen as to help the training process of the learning agent.

TTL is interpreted over finite traces $\lambda$, where $|\lambda|$ denotes the length of the trace. We denote time steps, i.e., instants, on the trace as $\lambda[j]$, for $0 \leq j < |\lambda|$; whereas $\lambda[i,j]$ is the (sub)trace between instants $i$ and $j$.
In order to define the satisfaction relation for TTL, we associate with every atomic task $\alpha$ an atomic proposition $p_{\alpha}$, which represents $\alpha$'s fulfilment condition. Note again that, in this context, $\alpha$ is a {\em reachability goal}, typically expressed in TL as an {\em eventuality} $\diamond p_{\alpha}$.
Then, a {\em model} is a tuple $ \mathcal{N} = \langle \mathcal{M}, L \rangle $, where $\mathcal{M}$ is a PONMRDP, and $L : S \to 2^{AP}$ is a labelling of states in $S$ with truth evaluations of atoms $p_{\alpha}$ in some set $AP$ of atoms. 
\begin{definition}[Satisfaction] \label{satisfaction}
Let $ \mathcal{N}$ be a model and $\lambda$ a finite trace. We define the satisfaction
relation $\models$ for tasks $T$ and formulae $\phi$ on trace $\lambda$ inductively as follows:
\begin{tabbing}
$( \mathcal{N}, \lambda) \models T \cup T'$ \ \  \= iff \ \ \= for some $0 \leq j < |\lambda|$, $( \mathcal{N}, \lambda) \models T$ or $( \mathcal{N}, \lambda) \models T'$\kill
$(\mathcal{N}, \lambda) \models \alpha$ \> iff \> for some $0 \leq j < |\lambda|$, $p_{\alpha} \in L(\lambda[j])$\\
$(\mathcal{N}, \lambda) \models \alpha \sim$ \> iff \> for some $0 \leq j < |\lambda|$, for some $q \neq$ \\ 
\> \> $ p_{\alpha}$, $q \in L(\lambda[j])$ and $p_{\alpha} \notin L(\lambda[j])$ \\
$( \mathcal{N}, \lambda) \models \alpha \lor \alpha'$ \> iff \> $( \mathcal{N}, \lambda) \models \alpha$ or $( \mathcal{N}, \lambda) \models \alpha'$\\
$( \mathcal{N}, \lambda) \models \phi ; \phi'$ \> iff \> for some $0 \leq j < |\lambda|$,  $( \mathcal{N},\lambda[0,j])$ \\
\> \> $\models \phi$ and $( \mathcal{N}, \lambda[j+1,|\lambda-1|]) \models \phi'$\\ 
$( \mathcal{N}, \lambda) \models \phi \cup \phi'$ \> iff \> $( \mathcal{N}, \lambda) \models \phi$ or $( \mathcal{N}, \lambda) \models \phi'$
\end{tabbing}
\end{definition}

\looseness=-1
By Def.~\ref{satisfaction}, an atomic task $\alpha$ indeed corresponds to a formula $\diamond p_{\alpha}$ of temporal logic, where $p_{\alpha}$ is the fulfilment condition associated with $\alpha$. It can be immediately inferred that TTL is a fragment of the widely-used Linear-time Temporal Logic over finite traces (LTL$_f$) \cite{de2013linear}. We provide a translation of TTL into LTL$_f$ and the corresponding proof of truth preservation in Sec.~\ref{subsec:AppLTLf} in the supplemental material. To better understand the language we present a toy example.
 
\begin{example}\label{example:Succ_Exec}
The specification ``get something different from grass, then collect grass and later use either the workbench or the toolshed" can be expressed in TTL as:
    $$\phi \triangleq (grass\sim) ; grass ; (workbench \cup toolshed)$$
\end{example}
  
\looseness=-1

TTL is designed to be expressive enough to encapsulate both the tasks described in \cite{andreas2017modular}, a popular benchmark in the RL-TL literature \cite{leon2020extended,toro2018teaching, de2019foundations}, and the negated instructions from studies about negation in natural language \cite{hill2020environmental}, i.e., encapsulating ``not wood" as ``get something different from wood".
\begin{rem*}
As a foundational work assessing the ability of learning agents to generalise abstract operators with unseen instructions, we intentionally restricted TTL syntax with atomic operators to negation and disjunction. This allows to carefully study the ability of DRL agents to generalise with these two important operators.  
\end{rem*}


\subsection{The Symbolic module} \label{subsec:Smodule}

\input{SymbModule}

\subsection{The Neural module} \label{subsec:NeuralMod}

\input{NeurModule}

%% file: SymbModule.tex
Given an instruction $\phi$ in TTL, the first task of the symbolic module (SM), detailed in Algorithm~\ref{alg:SymbM}, is to decompose the instruction into lists of atomic tasks for the neural module (NM). Intuitively, the SM transforms a PONMRDP into a progression of POMDPs, which is a well-known procedure in the literature on RL with TL specifications \cite{brafman2018ltlf}.
 \input{utils/algorithms/SM}
\looseness=-1
In particular, a task extractor $\mathcal{E}$ transforms $\phi$ into the set $\mathcal{K}$ of all sequences of tasks $T$ that satisfy $\phi$. As standard in the literature \cite{andreas2017modular, kuo2020encoding}, we assume that the SM has access to an internal labelling function $\mathcal{L}_I: Z \rightarrow 2^{AP}$, which maps the agent's observations into sets of atoms in $AP$. Note that this $\mathcal{L}_I$ might differ from the labelling $\mathcal{L}$ originally defined in Sec.~\ref{subsec:TTL}, since $\mathcal{L}$ maps states to sets of atoms. For our approach to be effective, observations have to contain enough information to decide the truth of atoms, so that both the agent's internal and the model's overall labelling functions remain aligned. Given $\mathcal{L}_I$ and the sequences of tasks in $\mathcal{K}$, a progression function $\mathcal{P}$ selects the next task from $\mathcal{K}$ that the NM has to solve. Once a task is solved, $\mathcal{P}$ updates $\mathcal{K}$ and selects the next task. This process is repeated until $\phi$ is satisfied or the time limit is reached. We include the pseudocode for $\mathcal{P}$ and $\mathcal{K}$ subroutines in Appendix~\ref{subsec:App_SM}.


In order to transmit the current task $T$ to the NM, the SM expands observation $z$ with the TTL formula of $T$. In the current context $z$ is a square visual input with a fixed perspective. This is expanded either by adding extra rows to $z$ containing $T$ (see Figure~\ref{fig:randmaps} center) or providing $T$ in a separated "language" channel (see Figure~\ref{fig:randmaps} right). During the episode the 
SM also generates an internal reward signal that guides the NM to solve $T$:

\begin{equation*}
\small
R_I\left(p_t\right) =
\begin{cases}
$d$ & \text {if } p_t=\emptyset, \text { where $d < 0$};\\
$i$ & \text {if } p_t=p_\alpha, \text { for } \alpha \text{ occurring in } T \text { and } \\
&   T \neq \alpha \sim \text {, or } \text {if } p_t \neq p_\alpha, \text { for } \alpha \\
&   \text{ occurring in } T \text { and } T \equiv \alpha \sim, \text { where } i \geq 0 ;\\
$c$ & \text {otherwise, where $c << d$. }
\end{cases}
\end{equation*}
where $\mathcal{L}_I(z_t)=p_t$ for any time step $t$. Intuitively, the NM is rewarded when an event detector ($\mathcal{L}_I$) fires a signal that satisfies the current task, and penalised when $\mathcal{L}_I$ signals an event  not related to $T$. A small penalisation is also given at each time step to induce the agent to promptly solve $T$.  

 We now illustrate the inner working of the SM: 
 
\looseness=-1
\begin{example}

Consider one of the complex specifications used in Sec.~\ref{sec:experiments}: 
$\phi_1= ((wood ; grass) \cup (iron; axe));workbench; $ \\ $ (toolshed \sim)$. Given $\phi_1$, the task extractor $\mathcal{E}$  outputs set $\mathcal{K} = \{ \langle [wood], [grass], [workbench],[toolshed \sim] \rangle ,\langle [iron], $ \\$ [axe],  [workbench], [toolshed \sim] \rangle \}$ with two available sequences of tasks. Set $\mathcal{K}$ is forwarded to the progression function $\mathcal{P}$ and since we have two lists with different initial elements, both positive, $\mathcal{P}$ selects $[wood \lor iron]$ as task  $T$ to extend the next observations. The NM iterates with the environment using the extended observations as input. At each iteration, functions $\mathcal{L}_I$ and $R_I$ respectively evaluate the progression of  $T$ and reward the NM until  $T$ is satisfied. In our example, if the NM got $wood$, the second sequence and $[wood]$ are discarded and the next task becomes $[grass]$. This process is repeated until $\mathcal{K}$ is empty or the time limit is reached.
\end{example}

%% file: utils/algorithms/SM.tex
\begin{algorithm}[t]
\caption{The symbolic module (SM)}
\label{alg:SymbM}
\begin{algorithmic}[1]
\STATE {\bfseries Input:} Instruction $\phi$
    \STATE Generate the accepted sequences of tasks $\mathcal{K} \leftarrow \mathcal{E}(\phi)$
    \STATE Retrieve the current observation: $z$
    \STATE Get the true proposition: $p \leftarrow \mathcal{L}_I(z)$
    \STATE Get the first task:  $T \leftarrow \mathcal{P}$($\mathcal{K}, p$)
     \REPEAT
         \STATE Generate the extended observation: $z^{ext} \leftarrow z +  T$
         \STATE Get the next action: $a \leftarrow \text{NM}(z^{ext})$
         \STATE Execute $a$, new observation $z'$
          \STATE New true proposition: $p' \leftarrow \mathcal{L}_I(z')$
          \STATE Provide the internal reward: NM $\leftarrow R_I(p')$
          \IF{$p' == p_\alpha$ for $\alpha \in T$}
            \STATE Update $\mathcal{K}, T' \leftarrow \mathcal{P}$($\mathcal{K}, p'$)
          \ENDIF
    \UNTIL{$\mathcal{K} == \emptyset$ or time limit}
\end{algorithmic}
\end{algorithm}

%% file: NeurModule.tex
\looseness=-1
The neural module consists of a DRL algorithm that learns to execute TTL tasks. We use A2C, a synchronous version of the algorithm introduced by \citet{mnih2016asynchronous}. The NM is 
trained in procedurally generated maps, using a fixed perspective which is known to help agents with generalisation \cite{mao2019neuro}. All the 
agents share the same hyperparameters (detailed in Appendix~\ref{subsec:App_ExpDet}) but differ on the CNN architecture.

\looseness=-1
\paragraph{CNN architecture.}  We study architectures with convolutional layers followed by a fully connected (f.c.) layer \cite{goodfellow2016deep} and a recurrent layer (an LSTM from \citet{gers2000learning}). In preliminary studies, we observed that agents with different f.c. and recurrent layer configurations can exhibit similar generalisation abilities. However, as detailed in Sec.~\ref{sec:experiments}, different convolutional configurations yield significantly different results. Particularly, while all the convolutional architectures we test achieve a similarly good performance with training instructions, only specific configurations correctly execute zero-shot tasks. We find that the ability to generalise is correlated to the degree of {\em alignment} between the CNN architecture and the environment. In our gridworld settings, we say that a CNN is weakly aligned (WA) with an environment when the kernel dimensions and stride size are factors of the tile resolution. If a CNN is WA and the two first dimensions (length and width) of its visual embedding correspond to the number of input tiles in the original observation, we say that this CNN is strongly aligned (SA). Note that the number of channels of the output, i.e., the third dimension of the embedding, does not influence the alignment. Last, when a CNN is not WA or SA, we say it is not aligned (NA). When aiming systematic learning from $\sim$6 training instructions, we find that only SA networks (when giving visual instructions) or SA and WA networks (when instructions are given in a separated channel) correctly execute zero-shot tasks.

%% file: experiments.tex
\begin{figure*}[t]
    \includegraphics[width=0.30\textwidth, height=4.8cm]{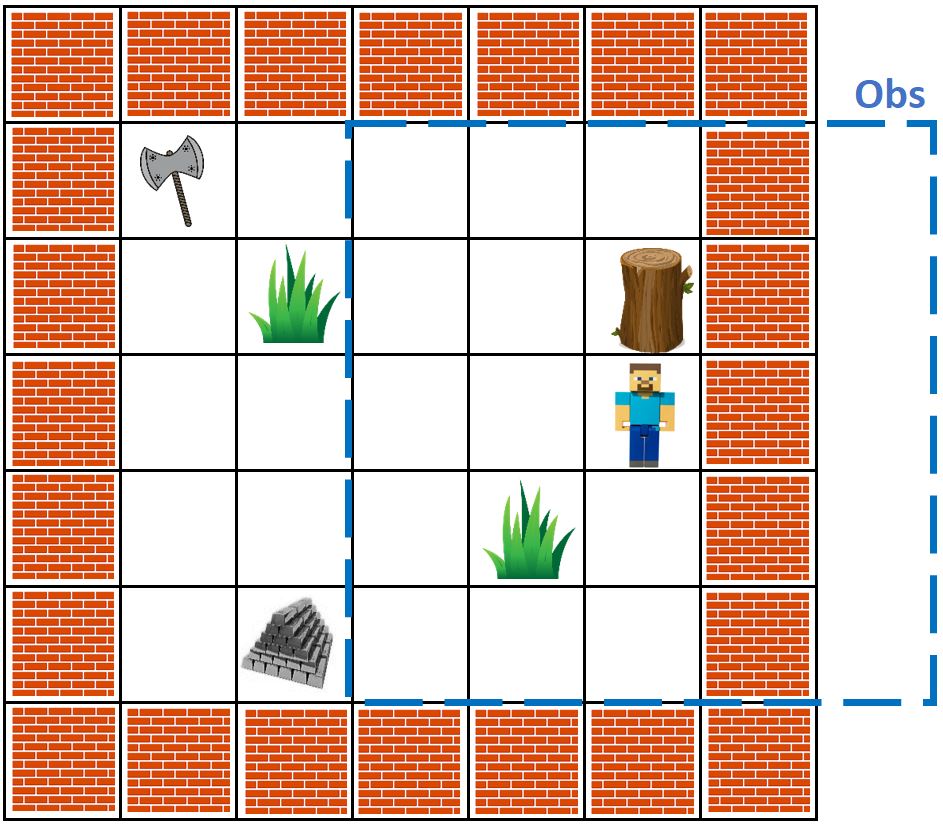}
    \includegraphics[width=0.28\textwidth, height=4.5cm]{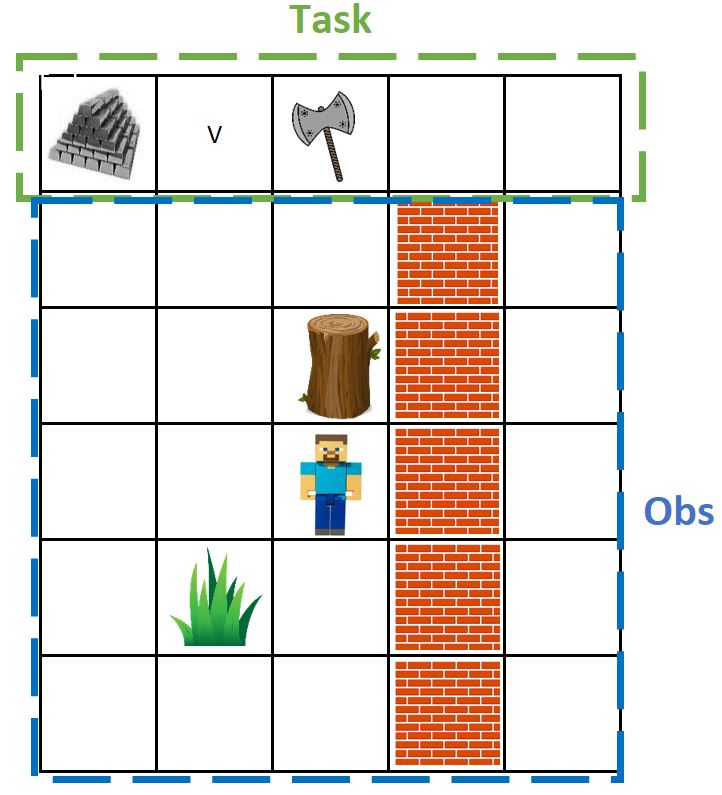}
    \hfill
    \includegraphics[width=0.28\textwidth, height=5cm]{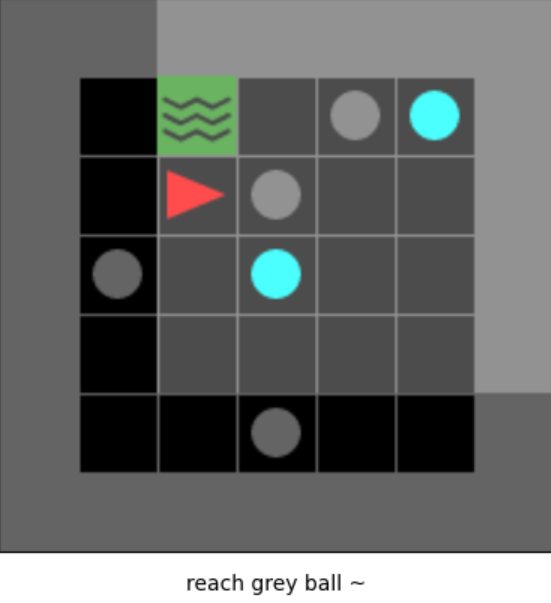}
    \caption{\textbf{Left}. Example of a procedurally generated test map (left) in the Minecraftwith its corresponding extended observation (\textbf{center}) when the agent's task is to get either iron or an axe (top row). Tasks are specified in the top row of the extended observation in this setting by depicting the object itself. \textbf{Right}. A MiniGrid map where the task is to reach an object different from the grey ball. Instructions are given on a separated channel. Here objects are combinations of shapes and colors.}  \label{fig:randmaps}
\end{figure*}

We evaluate our framework in procedurally generated grid-worlds. The first setting (Figure~\ref{fig:randmaps} left) is designed along the lines of a Minecraft-inspired (Minecraft for sort) benchmark widely used in RL-TL \cite{andreas2017modular, toro2018teaching, de2019foundations}, which we adapt to match the 2D maps from \citet{hill2020environmental} (the last study about negation). The action set consists of 4 individual actions: {\em Up, Down, Left, Right}, that move the agent one tile in the given direction. In this setting each tile has a resolution of 9x9x1 values. The DRL algorithm receives the observation extended with visual instructions depicting the target objects (Figure~\ref{fig:randmaps} center). We also evaluate the agents in the popular MiniGrid benchmark \cite{gym_minigrid}, depicted in Figure~\ref{fig:randmaps} right. Here each tile has a resolution of 8x8x3 values, and the action set is {\em Move forward, Turn left, Turn right}. In this benchmark instructions are given through a separated language channel.

For each setting we test four different CNN configurations. The first two are common architectures from the literature: {\em Atari-CNN} \cite{mnih2015human, mnih2016asynchronous} which is NA with Minecraft and WA with MiniGrid, and {\em ResNet} \cite{espeholt2018impala, hill2020environmental} WA with Minecraft and NA with MiniGrid. We contrast these networks with different architectures designed to be SA with each environment and that we refer to as Env-CNN. Last, for each environment we also test the performance of the same Env-CNNs, but when the c.l. are frozen during training, i.e., the c.l. use randomly initialized features. We refer to these networks as Env-CNN\textsuperscript{RF}. Schemes of the architectures are given in Appendix~\ref{app: subsec: arch}.

In every setting we refer to the global set of objects as $\mathcal{X}$. $\mathcal{X}$ is split into training ($\mathcal{X}_1$), validation ($\mathcal{X}_2$) and test sets ($\mathcal{X}_3$) where  $\mathcal{X}= \mathcal{X}_1 \cup \mathcal{X}_2 \cup \mathcal{X}_3$, $\mathcal{X}_1 \cap \mathcal{X}_2 = \emptyset$, $\mathcal{X}_1 \cap \mathcal{X}_3 = \emptyset$, and $\mathcal{X}_2 \cap \mathcal{X}_3 = \emptyset$. We contrast the results of learning with training sets of different size that we refer to as {\em small} ($|\mathcal{X}_{1}|=6$) and {\em large} ($|\mathcal{X}_{1}|=20$). In Minecraft, agents are trained with objects from $\mathcal{X}_1$ only. In MiniGrid, where agents require to ground words to objects, we train the agents with positive instructions from $\mathcal{X}$, while instructions concerning the negation or disjunction operators refer only to objects in $\mathcal{X}_1$. Agents are trained with atomic tasks $T$. At each episode the agent navigates receiving rewards according to $T$. The episode ends when the agent receives a positive reward (meaning that $T$ is satisfied) or the time limit is reached. During training we periodically evaluate the performance of the agents with the coresponding validation set $\mathcal{X}_2$, that we use to select the best learning weights in each run. Further details about object set generation, rewards, environment mechanics and training and validation plots are included in Appendix~\ref{subsec:App_ExpDet}.

\input{utils/table1}

\input{utils/table2}

\looseness=-1
\paragraph{Results about systematic learning.} Table~\ref{table:tasks} presents the raw reward results in $\mathcal{X}_1$ and $\mathcal{X}_3$ with the different CNNs. Note that ``train" refers to results with objects from  $\mathcal{X}_1$ but with the weights that achieve the best rewards in $\mathcal{X}_2$. Consequently, all the networks can accumulate higher rewards in training if we let them overfit. Results in Table~\ref{table:tasks} alone do not yield a strong correlation between the alignment of the CNN and the generalisation ability. Yet, as anticipated in Sec.~\ref{sec:backg} this correlation becomes evident when evaluating whether the networks are actually following the test instructions. Table~\ref{table:BCMs} shows the results in a particular set that we call binary choice maps (BCMs). Here maps have only two objects (except in ``disjunction 2 choices", with four), one granting a positive reward ($i=+1$) the other a penalisation ($c=-1$). In ``disjunction 2 choices" two objects penalise, and two provide a reward. Since agents are always trained with reliable instructions pointing to objects that grant a positive reward, an agent that is correctly generalising to zero-shot instructions should collect significantly higher rewards with reliable instructions than with deceptive ones (i.e., the later instructions point to objects that penalise). Roughly, agents should accumulate with reliable instructions two times greater rewards than when given deceptive instructions to show similar generalisation to the experiments in previous studies about negation with 100 training instructions \cite{hill2020environmental}. 

\looseness=-1
The first strong pattern that we extract from Table~\ref{table:BCMs} is that only architectures with some form of alignment (WA and SA) give evidence of systematic learning when no more than six training instructions are available. NA architectures show minimal to none generalisation (e.g. better performance with deceptive instructions than with reliable) in all the small settings. With 20 objects, NA architectures also fail when instructions are visual (Minecraft) while struggling with the disjunction operator when instructions come in a dedicated channel (minigrid). In the latter setting, architectures with some alignment (WA) show good generalisation performance. Yet, when instructions are visual they only learn to generalise positive and negative instructions, and exclusively in the large setting. Notably, architectures that are strongly aligned (SA) show good generalisation even when the weights of the c.l. are not trained (Env-CNN\textsuperscript{RF}), further evidencing the high influence of the architecture in how the agent generalises. Remarkably, the Env-CNN (the fully trained SA architecture) is the only configuration that systematically achieves a performance greater than 1 (the performance of a random agent) with reliable instructions while getting less than 1 with deceptive instructions across all the settings. That is, {\em SA architectures are the only configurations that have systematically learnt to generalise from the training instructions to guess not only the desired object/s, but also which undesired items should be avoided according to the test instruction}.

\looseness=-1
Last, we highlight that in preliminary studies we also explored additional overfitting-prevention methods, including dropout and autoencoders, to help NA networks. However, the results do not vary from the ones obtained by the train-validation-test set division that we carry here and that evidence the benefits of using architectures that are aligned with its environment. We only include the disjunction test where agents struggled the most, due to space constrains, additional results are included in Appendix~\ref{sec:App_Disj}.

\input{utils/table5}

\paragraph{Generalisation with TTL formulae.} We last evaluate the performance of the same agents trained with atomic tasks $T$ when executing complex instructions given as TTL formulae $\phi$ that refer to zero-shot objects. Table~\ref{table:ComplexT} shows the average reward and number of actions needed in 200 maps executing complex instructions from Appendix~\ref{app: subsec: complex}. In line with the results with TTL tasks from Table~\ref{table:tasks}, we see that a stronger alignment yields better the final performance (i.e., SA$>$WA$>$NA accumulated reward).

%% file: utils/table1.tex
\begin{table*}[t]
    \centering
  \mycfs{8.2}
   \begin{tabular}{ccccccccc}
    \toprule
\textit{Minecraft}  &\multicolumn{2}{c}{Atari-CNN (\textbf{NA})} &  \multicolumn{2}{c}{ResNet (\textbf{WA})}  & \multicolumn{2}{c}{Env-CNN\textsubscript{minecraft} (\textbf{SA})} 
  & \multicolumn{2}{c}{Env-CNN\SPSB{RF}{minecraft} (\textbf{SA})}\\
\cmidrule(r){2-3}
\cmidrule(r){4-5}
\cmidrule(r){6-7}
\cmidrule(r){8-9}
  $|\mathcal{X}_{1}|$ & Train & Test & Train & Test & Train & Test &  Train & Test  \\
  6  & 3.04$(0.12)$ &	\textbf{2.87}$(0.21)$ &	3.24$(0.49)$ & 2.72 $(0.22)$ & \textbf{4.46$(0.75)$} & 1.8$(0.25)$ & 2.90$(0.81)$ & 2.58$(0.76)$\\
  
 20 & 3.11$(0.14)$ & 3.06$(0.08)$	& \textbf{4.52$(0.44)$} & 2.14$(0.50)$	& 4.45$(0.30)$	& \textbf{3.81$(0.72)$}& 3.28$(0.98)$ & 3.10$(0.88)$ \\

 \midrule    

\textit{Minigrid}  &\multicolumn{2}{c}{Atari-CNN (\textbf{WA})} &  \multicolumn{2}{c}{ResNet (\textbf{NA})}  & \multicolumn{2}{c}{Env-CNN\textsubscript{minigrid} (\textbf{SA})}  & \multicolumn{2}{c}{Env-CNN\SPSB{RF}{minigrid} (\textbf{SA})}\\
\cmidrule(r){2-3}
\cmidrule(r){4-5}
\cmidrule(r){6-7}
\cmidrule(r){8-9}
 $|\mathcal{X}_{1}|$ & Train & Test & Train & Test & Train & Test &  Train & Test  \\
   6  & \textbf{4.02}$(1.09)$ &	\textbf{3.96}$(0.85)$ &	3.98$(0.93)$ & 3.55 $(0.69)$ & 2.75$(1.03)$ & 2.29$(1.03)$ & 3.31$(0.84)$ & 3.39$(0.31)$\\
  
 20 & 6.19$(0.75)$ & 3.33$(1.06)$	& 5.61$(0.78)$ & \textbf{5.60}$(0.99)$	& \textbf{7.31$(0.50)$}	& 4.09$(0.92)$&	4.74$(0.15)$ & 2.55$(3.39)$ \\

    \bottomrule
  \end{tabular}
   \normalsize
    \caption{Mean and standard deviation from the average rewards obtained from five independent runs in a set of 500 maps with atomic tasks. A value of 1.0 is the performance of a random walker. The highest value over all networks is bolded.} \label{table:tasks}
\end{table*}

%% file: utils/table2.tex
\begin{table*}[t]
\renewcommand{\arraystretch}{1}
  \centering
    \mycfs{8.2}
   \begin{tabular}{m{1.6cm}P{0.8cm}cccccccc}
    \toprule
\textit{Minecraft} &  $|\mathcal{X}_{1}|$ &\multicolumn{2}{c}{Atari-CNN (\textbf{NA})} &  \multicolumn{2}{c}{ResNet (\textbf{WA})}  & \multicolumn{2}{c}{Env-CNN\textsubscript{minecraft} (\textbf{SA})} 
  & \multicolumn{2}{c}{Env-CNN\SPSB{RF}{minecraft} (\textbf{SA})}\\
\cmidrule(r){3-4}
\cmidrule(r){5-6}
\cmidrule(r){7-8}
\cmidrule(r){9-10}
Instruction  & & Reliable & Deceptive & Reliable & Deceptive & Reliable & Deceptive &  Reliable & Deceptive  \\
\cmidrule(r){1-2}
 Positive & 6  & 3.89 $(0.12)$ & 3.92$(0.09)$
        &	3.72$(0.49)$ & 3.42$(0.81)$ & \textbf{3.25$(1.89)$} & \textbf{0.26$(1.88)$} & 
        \textbf{3.95}$(1.8)$ &\textbf{0.74$(0.65)$}\\

  & 20 & 4.14$(0.07)$ & 4.16$(0.09)$ &     
    \textbf{3.30$(1.83)$} &\textbf{ 1.10$(1.38)$}	& \textbf{5.43}$(0.24)$ &	\textbf{0.11}$(0.03)$  &  
    \textbf{3.95$(1.76)$} &\textbf{ 1.10$(1.18)$}\\

\cmidrule(r){1-2}
Negative & 6  & 3.82 $(0.18)$ & 3.81$(0.10)$
        &	3.52$(0.52)$ & 3.49$(0.68)$ & \textbf{2.20$(1.30)$} & \textbf{0.44$(0.38)$} & 
        \textbf{2.23}$(1.89)$ &\textbf{1.04$(0.82)$}\\

  & 20 & 3.94$(0.13)$ & 3.97$(0.09)$&     
   \textbf{ 3.09$(1.73)$} & \textbf{1.26$(0.93)$}	& \textbf{4.40}$(0.66)$ &	\textbf{0.14}$(0.05)$  &  
   \textbf{ 2.25$(1.80)$} & \textbf{0.94$(1.15)$}\\
\cmidrule(r){1-2}    
Disjunction & 6  & 2.56 $(0.23)$ & 2.59$(0.57)$
        &	2.29$(0.12)$ & 2.51$(0.69)$ & \textbf{2.23$(4.79)$} & \textbf{0.57$(0.29)$} & 
        2.54$(0.67)$ &1.37$(0.06)$\\

 2 choices & 20 & 2.63 $(0.23)$ & 2.68$(0.57)$
        &	2.16$(0.12)$ & 2.23$(0.69)$ & \textbf{3.53$(4.79)$} & \textbf{0.44$(0.29)$} & 
        \textbf{2.82$(0.67)$ }& \textbf{1.07$(0.90)$}\\\\
    
\midrule    

\textit{Minigrid} &  $|\mathcal{X}_{1}|$ &\multicolumn{2}{c}{Atari-CNN (\textbf{WA})} &  \multicolumn{2}{c}{ResNet (\textbf{NA})}  & \multicolumn{2}{c}{Env-CNN\textsubscript{minigrid} (\textbf{SA})} 
  & \multicolumn{2}{c}{Env-CNN\SPSB{RF}{minigrid} (\textbf{SA})}\\
\cmidrule(r){3-4}
\cmidrule(r){5-6}
\cmidrule(r){7-8}
\cmidrule(r){9-10}
Instruction   & & Reliable & Deceptive & Reliable & Deceptive & Reliable & Deceptive &  Reliable & Deceptive  \\
\cmidrule(r){1-2}
Negative & 6  & \textbf{2.83} $(0.96)$ & \textbf{0.44}$(0.25)$
        &	3.02$(1.37)$ & 2.97$(1.39)$ & \textbf{2.01$(1.27)$} & \textbf{0.41$(0.25)$ }& 
        \textbf{1.61}$(0.21)$ &\textbf{0.75$(0.16)$}\\

  & 20 & \textbf{2.77}$(0.79)$ &\textbf{ 0.34}$(0.10)$&     
    \textbf{5.05$(1.27)$ } & \textbf{0.20$(0.10)$}	& \textbf{5.03}$(1.65)$ & \textbf{0.16}$(0.05)$  &  
    \textbf{2.16$(0.23)$} & \textbf{0.98$(0.22)$}\\

\cmidrule(r){1-2}
Disjunction & 6  & \textbf{4.21} $(0.46)$ & \textbf{0.84} $(0.50)$
        &	4.02$(0.39)$ & 3.68$(0.59)$ & \textbf{2.67} $(1.49)$ & \textbf{ 0.79}$(0.54)$ & 
        \textbf{4.06}$(0.74)$ &\textbf{0.63$(0.44)$}\\

2 choices  & 20 & \textbf{3.42}$(0.97)$ & \textbf{1.14}$(0.48)$&     
    \textbf{5.03$(0.59)$} & \textbf{ 1.75$(0.77)$}	& \textbf{3.13}$(0.47)$ &	\textbf{0.37}$(0.09)$  &  
    2.72$(0.97)$ &1.58$(0.38)$\\
\cmidrule(r){1-2}    
Disjunction & 6  & \textbf{2.83 $(0.93)$} & \textbf{0.90}$(0.41)$
        &	3.52$(0.79)$ & 3.86$(1.30)$ &\textbf{ 1.68$(1.25)$} & \textbf{0.75$(0.27)$} & 
        \textbf{2.81}$(0.79)$ &\textbf{0.88$(0.55)$}\\

 2\textsuperscript{nd} choice & 20 & 2.11$(1.42)$ & 1.06$(0.32)$&     
    2.60$(0.74)$ & 2.37$(1.03)$	& \textbf{1.81}$(0.52)$ &	\textbf{0.50}$(0.08)$  &  
    1.91$(0.76)$ & 1.28$(0.4)$\\
    
 \bottomrule
\end{tabular}
\normalsize
\caption{Results from 500 binary choice maps with zero-shot instructions. In 'Reliable' higher is better while in 'Deceptive' lower is better. Agents show a stronger systematic generalisation the bigger the difference between respective reliable and deceptive instructions. Positive and negative labels refer to atomic positive or negative instruction respectively pointing to one of the two objects present. In disjunction 2\textsuperscript{nd} choice the instructions refers to two objects, but only the second is present in the map. Disjunction 2 choices points to two objects in a map with 4 items. Results are bolded where performance with reliable instructions is greater or equal than two times the respective performance with deceptive instructions.}\label{table:BCMs}

\end{table*}

%% file: utils/table5.tex
\begin{table*}

  \centering
  \mycfs{8.2}
\begin{tabular}{lcccccccccc} 
\toprule
& \multicolumn{2}{c}{Atari-CNN (\textbf{NA})} &  \multicolumn{2}{c}{ResNet (\textbf{WA})}  & \multicolumn{2}{c}{Env-CNN\textsubscript{minecraft} (\textbf{SA})} 
  & \multicolumn{2}{c}{Env-CNN\SPSB{RF}{minecraft} (\textbf{SA})}\\
\cmidrule(r){2-3}
\cmidrule(r){4-5}
\cmidrule(r){6-7}
\cmidrule(r){8-9}
$|\mathcal{X}_1|$& Reward & Steps & Reward & Steps & Reward & Steps &  Reward & Steps  \\
 6  & 2.45 $(0.25)$ & \textbf{26.28$(4.96)$}
        &	2.68$(0.65)$ & 49.26$(28.89)$ & \textbf{3.08$(1.49)$} & 49.1$(25.79)$ & 
        2.74$(1.78)$ &56.51$(22.87)$\\

20 & 2.99$(0.21)$ & \textbf{20.96$(1.50)$}&     
    3.49$(0.68)$ & 27.36$(5.20)$	& \textbf{7.45}$(3.36)$ &	32.84$(6.79)$  &  
   5.49$(1.21)$ &38.18$(9.66)$\\
\bottomrule
\end{tabular}
\normalsize
 \caption{Results from a set of 200 maps with zero-shot complex instructions, i.e, longer than training instructions composed by various zero-shot tasks. Steps refers to the number of actions per instruction needed by the agents (a random walker requires 110 steps in average). The highest value over all networks is bolded.}
 \label{table:ComplexT}
\end{table*}

%% file: relatedWork.tex
This work bridges the fields of systematic generalisation with formal methods in RL. Here we include the most relevant literature from each field to this research:

\paragraph{Systematic Generalisation.} Generalisation beyond training instructions has been widely studied in the RL literature. For instance, 
\citet{oh2017zero} presents a framework that relies on hierarchical RL and task decomposition of instructions in English to enhance generalisation in DRL. Later work from \citet{mao2019neuro} introduces a neuro-symbolic concept learner that jointly learns visual concepts, words, and semantic parsing of sentences from natural supervision. \citet{yu2018interactive} present a model that reports strong generalisation in a 2D environment that is aimed at language understanding for positive instructions. Closer to our line of work, there is growing literature focused on finding 
the right drivers, e.g., fixed perspectives and varied training data, enabling systematic learning from natural language instructions (see e.g., \citet{smolensky1988proper, bahdanau2018systematic,lake2019compositional}). Recent work \cite{hill2020environmental} suggests that systematic generalisation is not a binary question, but an {\em emergent} property of agents interacting with a situated environment \cite{mcclelland2010letting} and explores the drivers that enable agents generalise an abstract operator such as negation. We expand the state of this line of works by first, providing evidence that systematic learning can emerge with logic operators from as few examples as with positive instructions, second, giving evidence that the CNN architecture is a key driver --that previous research was oblivious about-- towards generalisation.

\paragraph{Reinforcement Learning and Temporal Logic.} Training autonomous agents to solve multi-task goals expressed in temporal logic is drawing growing attention from the research community. Examples include following instructions expressed as TL formulae by using logic progression and RL \cite{littman2017environment, andreas2017modular}, tackling
continuous-action environments \cite{yuan2019modular}, multi-agent systems \cite{leon2020extended} and studies on the applicability of different TL languages \cite{camacho2019ltl}. Complex
goals expressed as temporal formulae can also be easily decomposed by using other techniques, including finite state automata \cite{de2019foundations, icarte2018using} and progression \cite{toro2018teaching}. Consequently, those works focus on solving non-Markovian models by extending the state space in a minimal form (to minimally impact the state space) so that the RL agent learns from an equivalent Markovian model, where an optimal behavior is also optimal in the original system \cite{bacchus1996rewarding,brafman2018ltlf}. A drawback of the mentioned literature is their exclusive focus on learning to solve training tasks. For this reason \citet{kuo2020encoding} targets generalisation of temporal logic instructions with DRL at the expense of requiring an additional neural network for each new symbol or object encountered. Our work advances the state of the art by presenting a framework that can execute zero-shot TL formulae while relying on a single neural network.

%% file: Conclusions.tex
\looseness=-1
We present a framework that can execute unseen formal instructions while learning from a limited number of instructions. With respect to generalisation, an obvious limitation of our work is the use of regular gridworlds in our experiments. However, this does not lessen our contribution since previous closely-related work have struggled to provide evidence of generalisation with abstract operators in similar settings.  Another limitation is that –- in line with related work on generalisation \cite{lake2018generalization, bahdanau2018systematic, hill2020environmental} -- we do not give nomological explanations for the generalisation ability (or lack thereof) that we evidence. Yet, this out our scope, that is proving that generalisation can emerge with logical operators as early as with positive instructions while giving evidence of the important role of the convolutional layers. Our findings suggest that from as few as six different instructions agents with aligned architectures learn not only how to follow training commands, but also abstract information about how symbols and operators in the given language compose and how the combination of those symbols influences what the agent should do. This contrasts with \citet{hill2020environmental} -- the only previous study evidencing some generalisation with unseen negated instructions -- which suggests that $\sim$100 instructions are required with abstract operators. Using the same form of negation, we find that the CNN architecture plays a key role when learning from limited examples. Still, we observe that the ResNet used in \citeauthor{hill2020environmental} shows evidence of generalising negation with 20 instructions in both settings. This also contrasts with \citeauthor{hill2020environmental}, where the ResNet performed barely better than random when trained with 40 negated instructions. We find that this difference comes from our use of a validation set, not common in RL literature and consequently not present in \citeauthor{hill2020environmental}. We provide empirical evidence in Appendix~\ref{sec:App_HillNeg}.

\looseness=-1
With respect to formal methods and RL (FM-RL) literature, we use a language that is less expressive than the latest contributions in the field. This constraint was needed to carry a careful experimental evaluation of the systematic learning with the different types of instructions. Given the positive results, more expressive languages can be used in future iterations. Nevertheless, we present the first framework that, leveraging on a learning-oriented syntax and the compositional structure of logical formulae, is capable of executing zero-shot complex formulae in temporal logic while relying on a single network. As anticipated in Sec.~\ref{sec:relatedWork}, this contrasts with dominant methods in the FM-RL literature that rely on a new policy network per task (e.g., \citet{de2019foundations, icarte2019learning}) or a new network per word \cite{kuo2020encoding}.

\looseness=-1
\paragraph{Concluding remarks.} We presented a foundational study that acts as a bridge between the areas of DRL, FM and systematic generalisation. Our framework demonstrates the possibility of generalise to unseen multi-task TL formulae, where future works may applying more sophisticated symbolic modules such as reward machines or automata. We have empirically proved that systematic learning can emerge with abstract operators such as negation from as few as 6 instructions. Hence, we plan exploring more expressive language without requiring large numbers of training instructions for new symbols. In the context of generalisation, it has been demonstrated the important role that convolutional layers play in how agents generalise. In our case, exploiting prior-knowledge of the environment -- the tile resolution of the gridworld -- was key to achieve better agents. Our findings suggest that when aiming generalisation in real-world environments, where prior-knowledge cannot be directly applied, it may be worth finding a suited CNN architecture rather than simply training a generic configuration. Literature on meta-learning and evolutionary techniques \cite{elsken2019neural, song2020rapidly} provide methods to automatically search well-suited architectures and could be beneficial for future works targeting generalisation.

%% file: appendix.tex
\pagebreak
\section{Experiment details} \label{subsec:App_ExpDet}
\input{Appendix/details}

\section{Symbolic Module} \label{subsec:App_SM}
\input{Appendix/SModule}
\section{Disjunction results} \label{sec:App_Disj}
\input{Appendix/control}
\section{The struggle with negated instructions} \label{sec:App_HillNeg}
\input{Appendix/StruggNegation}
\section{Requirements for Logical Operators}
\input{Appendix/operator}
\section{Translating TTL into LTL$_f$} \label{subsec:AppLTLf}

 It can be easily inferred that our task temporal logic (TTL) is a fragment of the more popular linear-time temporal logic over finite traces (LTL$_f$) \cite{baier2008principles}. The syntax of LTL$_f$ is defined as follows: 

\begin{equation*}
    \varphi \enskip::=\enskip p\mid\neg \varphi\mid \varphi_{1} \wedge \varphi_{2}\mid \bigcirc \varphi\mid \varphi_{1} \mathrm{U} \varphi_{2}
\end{equation*}

Note that both TTL and LTL$_f$ are defined over finite traces. Below we provide truth-preserving translations of TTL into LTL$_f$.

\begin{definition}
Translation $\tau$ from TTL to LTL$_f$ is defined as follows:
\small
 \begin{tabbing} 
      $\tau_1(T;T')$ \ \ \ \= $=$ \ \ \= $\diamond(\tau'(T) \wedge \tau(T'))$ \kill
     $\tau(\alpha)$ \> $=$ \> $\diamond p_{\alpha}$, where $p_{\alpha}$ is the fulfilment \\ \> \>condition associated with task $\alpha$ \\
      $\tau(\alpha \sim)$ \>  $=$ \> $\diamond (\bigvee_{\beta \neq \alpha} p_{\beta} \wedge \neg p_{\alpha})$ \\
      $\tau(\alpha \lor \alpha')$ \>  $=$ \> $\tau(\alpha) \lor \tau(\alpha')$ \\
      $\tau(T \cup T')$ \> $=$ \> $\tau(T) \lor \tau(T')$\\
      
      $\tau(T;T')$ \> $=$ \> $\begin{cases}
      \diamond (p_{\alpha} \land \tau(T)) & \text{if }  T = \alpha\\
      \diamond ((\bigvee_{\beta \neq \alpha} p_{\beta} \wedge \neg p_{\alpha}) \land \tau(T)) & \text{if }  T = \alpha \sim\\
        \diamond ((p_{\alpha} \lor p_{\alpha'}) \land \tau(T)) & \text{if }  T = \alpha \lor \alpha'\\
 \tau(T_1; (T_2;T')) & \text{if } T = T_1;T_2\\
\tau((T_1;T') \cup (T_2;T')) & \text{if } T = T_1 \cup T_2
\end{cases}$
 \end{tabbing}
\end{definition}

We immediately prove that translation $\tau$ preserve the interpretation of formulae in TTL.
\begin{prop} \label{prop1}
Given a model $\mathcal{N}$ and trace $\lambda$, for every formula $T$ in TTL, 
\begin{eqnarray*}
    (\mathcal{N}, \lambda) \models T & \text{iff} & (\mathcal{N}, \lambda) \models \tau(T)
\end{eqnarray*}
\end{prop}

In the main body of this work we stated that TTL is a fragment of LTL$_f$, and supported such claim by explaining that translations $\tau_1$ and $\tau_2$ preserve the interpretation of formulae in TTL. We include now a detailed proof for this claim:

\begin{prop}
Given a model $\mathcal{N}$ and trace $\lambda$, for every formula $T$ in TTL, 
\begin{eqnarray*}
    (\mathcal{N}, \lambda) \models T & \text{iff} & (\mathcal{N}, \lambda) \models \tau(T)
\end{eqnarray*}
\end{prop}

\begin{proof}
The proof is by induction on the structure of formula $T$. The base case follows immediately by the semantics of TTL and LTL$_f$.

As for the induction step, the case of interest is for formulae of type $T;T'$.
In particular, consider the case for $T = \alpha$. We then have that $(\mathcal{N}, \lambda) \models \alpha; T'$ iff for some $0 \leq j < |\lambda|$, $(\mathcal{N}, \lambda[0,j]) \models \alpha$ and $(\mathcal{N},\lambda[j+ 1,|\lambda]) \models T'$. By induction hypothesis, this is equivalent to $(\mathcal{N}, \lambda[0,j]) \models \diamond p_{\alpha}$ and $(\mathcal{N},\lambda[j+ 1,|\lambda]) \models \tau(T')$. Finally, this is equivalent to $(\mathcal{N}, \lambda) \models \diamond (p_{\alpha} \land \tau(T'))$. The other cases.
are dealt with similarly.

Finally, the case for $T \cup T'$ follows by induction hypothesis and the distributivity of $\diamond$ over $\lor$.
\end{proof}

%% file: Appendix/details.tex
Here we provide further details about the experimental setting and training procedure.
\subsection{Architecture configurations} \label{app: subsec: arch}
\input{utils/arquitectures}
Figure~\ref{fig:NNarchitectures} includes schemes of the neural network architectures from Section~\ref{sec:experiments}. On the left, we see the general setting for Minecraft, where tasks instructions are embedded within the observation, and MiniGrid, where instructions are given in a separated channel. Note that we intentionally use a significantly different number of channels in the last layers with Env-CNN\textsubscript{minecraft} and Env-CNN\textsubscript{minigrid}. We do this to evidence that the number of channels in the visual embedding does not affect the ability of SA architectures to generalise training instructions. In Sec~\ref{sec:experiments} we observed that both SA architectures show good generalisation performance in their respective settings.

\subsection{Environment}  \label{app: subsec: env}
\paragraph{Map generation} Training maps are procedurally-generated with a random number of objects that ranges from 2 to 8, there are no restrictions on the number of objects on a training map that can be the same as long as one of the objects solves the current task. As explained in the main text, the action set consists of 4 individual actions in Minecraft: {\em Up, Down, Left, Right}, that move the agent one slot from its current position according to the given direction; and 3 actions in MiniGrid: {\em Forward, Turn Left, Turn Right}. If the agent moves against a wall, its position remains the same. Note that moving against walls is discouraged by the internal reward function that provides a negative reward at each timestep if the agent has not solved the current task. 

\paragraph{Instruction generation.} Instructions are procedurally generated with objects from the corresponding set. For instance, with the small setting we have 6 different objects, meaning that agents are learning the negation operator from 6 training instructions. Consequently, the total number of training instructions in the small setting is 42 (6 positive, 6 negative and 30 disjunctions --we do not instruct disjunctions with the same object twice--).

\paragraph{Object-set generation.} In the Minecraft-inspired environment, we procedurally generate objects as a random matrix of 9x9 values. Small and large training sets use 6 to 20 procedural objects as described in the main text. For validation and test sets we use 6 objects respectively. In MiniGrid, objects are not procedural values but compositions of colors, e.g., red, and shapes, e.g., key. Here the small set uses 2 shapes and 3 colors while training with negative instructions and disjunction. In validation we use an additional shape combined with training colors. In test we use 2 new colors, the validation shape and a new test shape. In the large setting, training has 4 shapes and 5 colors, an additional shape with training colors for validation, and 2 new colors with a a test shape and the validation shape in test.

\paragraph{Rewards.} In both settings the agents received a penalisation of -1 when reaching an object that does not satisfy the task, and +1 when the opposite occurs. We explored different penalisation values per time-step to stop the agents from local-optima policies, e.g., not reaching any object, while still making beneficial avoiding undesired objects when possible. We did a grid-search in the range $[-0.01,-0.2]$ and achieved the best results with $-0.1$ in Minecraft and $-0.05$ in MiniGrid.

\subsection{Training hyperparameters and plots}
The A2C algorithm uses a learning rate of $8 e^{-5}$, this value for the learning rate is the one that give us the best results when doing grid search between $1 e^{-5}$ and $1 e^{-3}$, except for the ResNet that shows better results with the value of $2 e^{-5}$ that we use in the experiments. The entropy and value loss regularization terms are $1 e^{-3}$ (best results we found in the range [$1 e^{-4}$,0] and $0.5$ (as in \citet{mnih2016asynchronous}) respectively. The expected-reward discount factor is $0.99$, which is typical value for this term. Batch size is set to 512 values, which gives the best results in the range of ($[128,2048]$).

\begin{figure*}
    \centering \includegraphics[width=0.49\textwidth]{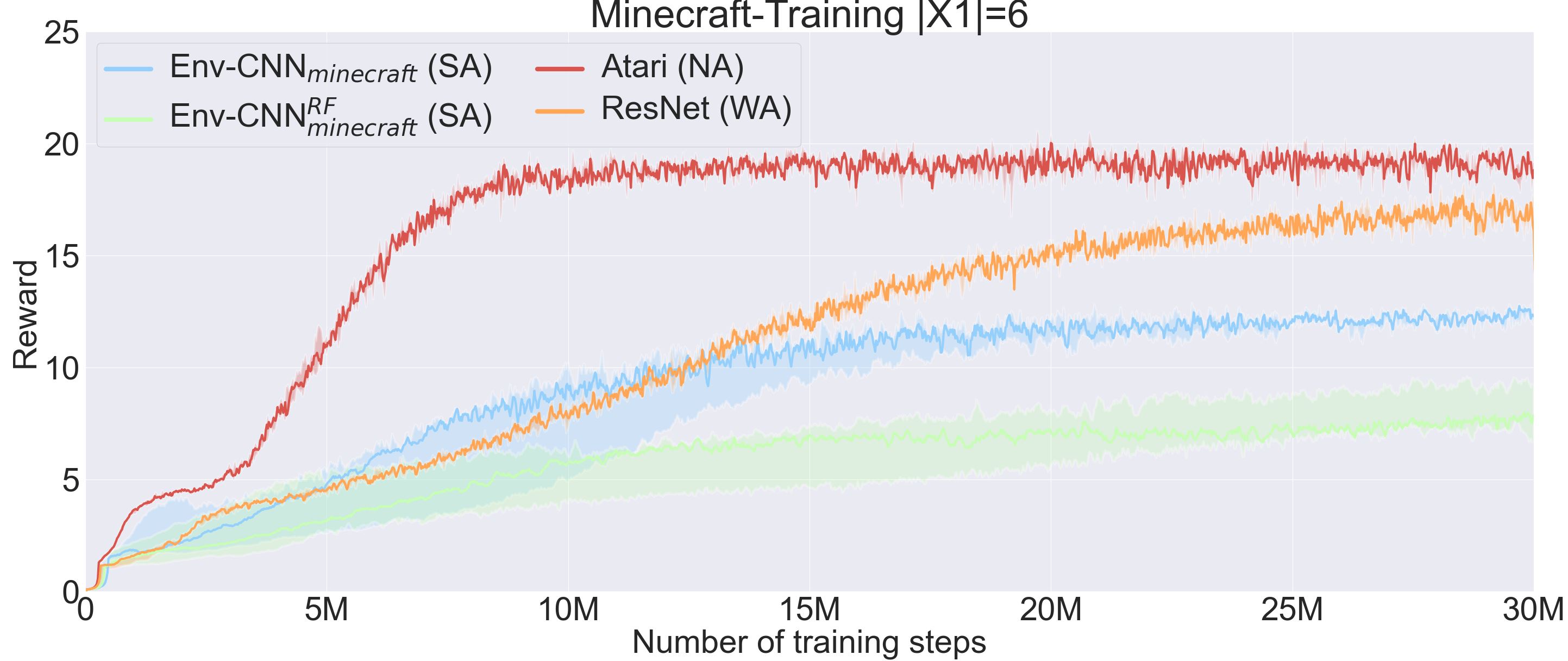}
    \centering \includegraphics[width=0.49\textwidth]{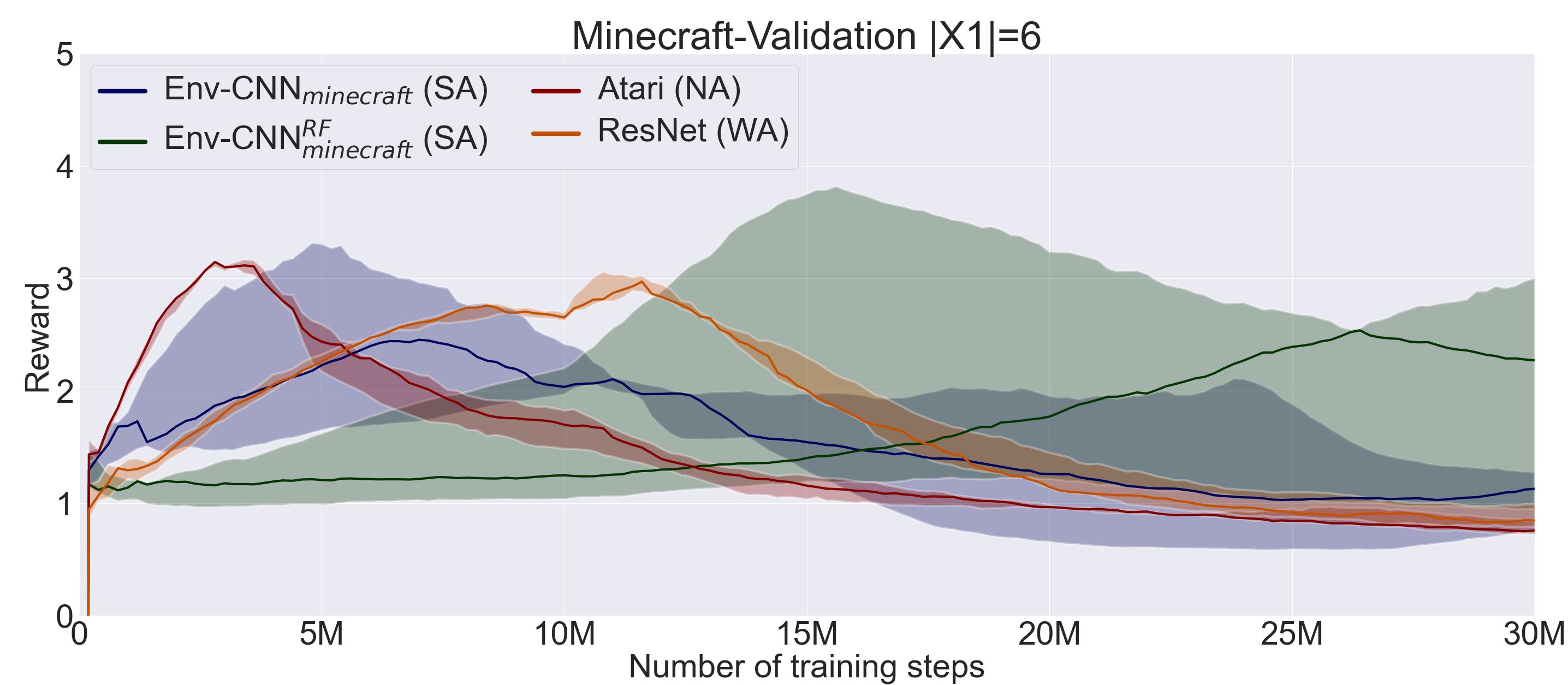}
    \vskip 0.15in
    \centering \includegraphics[width=0.49\textwidth]{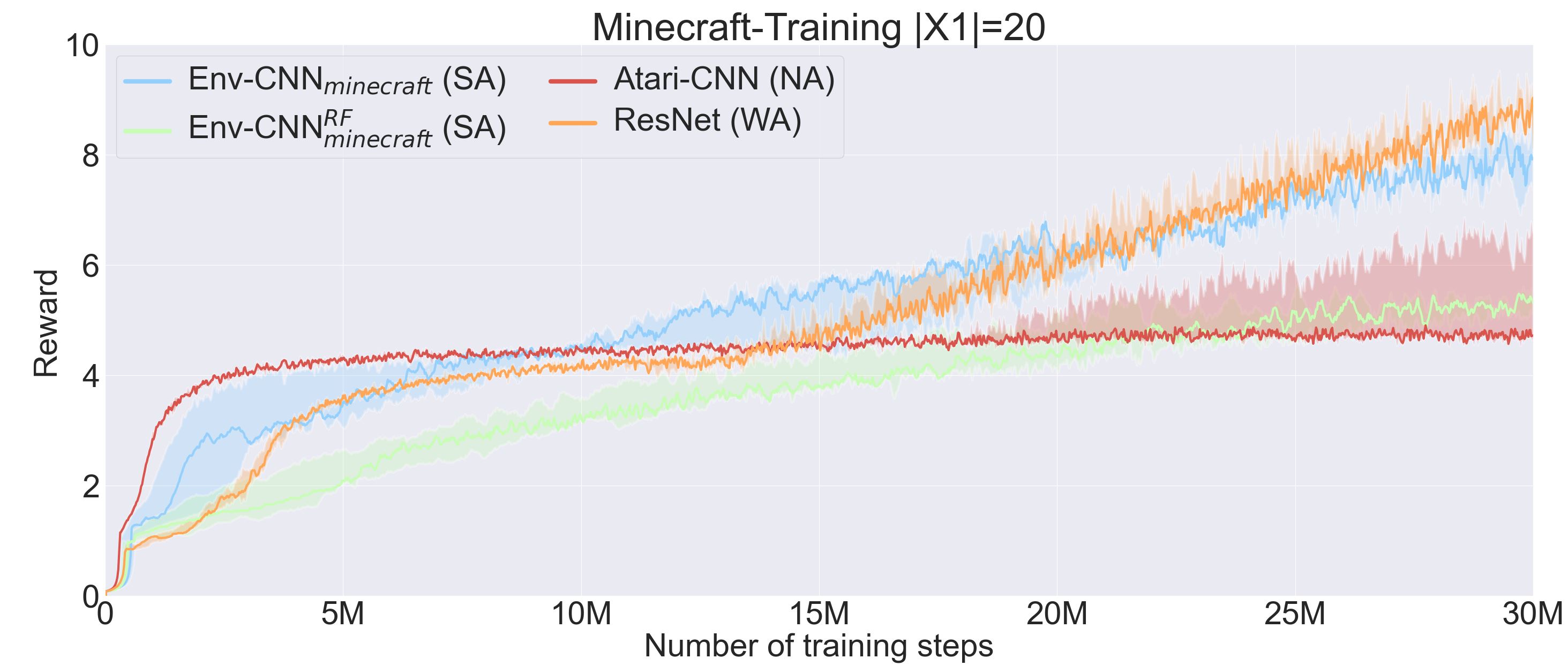}
    \centering \includegraphics[width=0.49\textwidth]{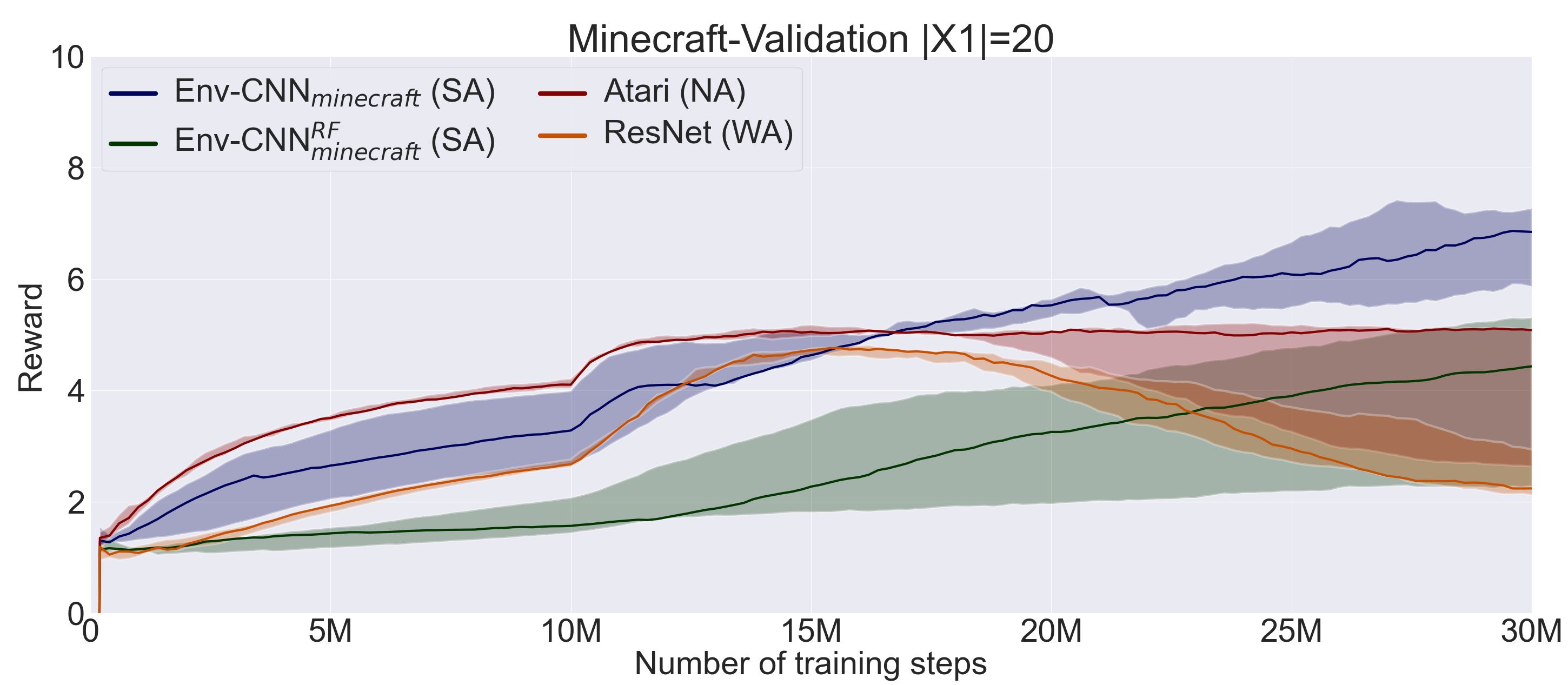}
    \vskip 0.15in
     \centering \includegraphics[width=0.49\textwidth]{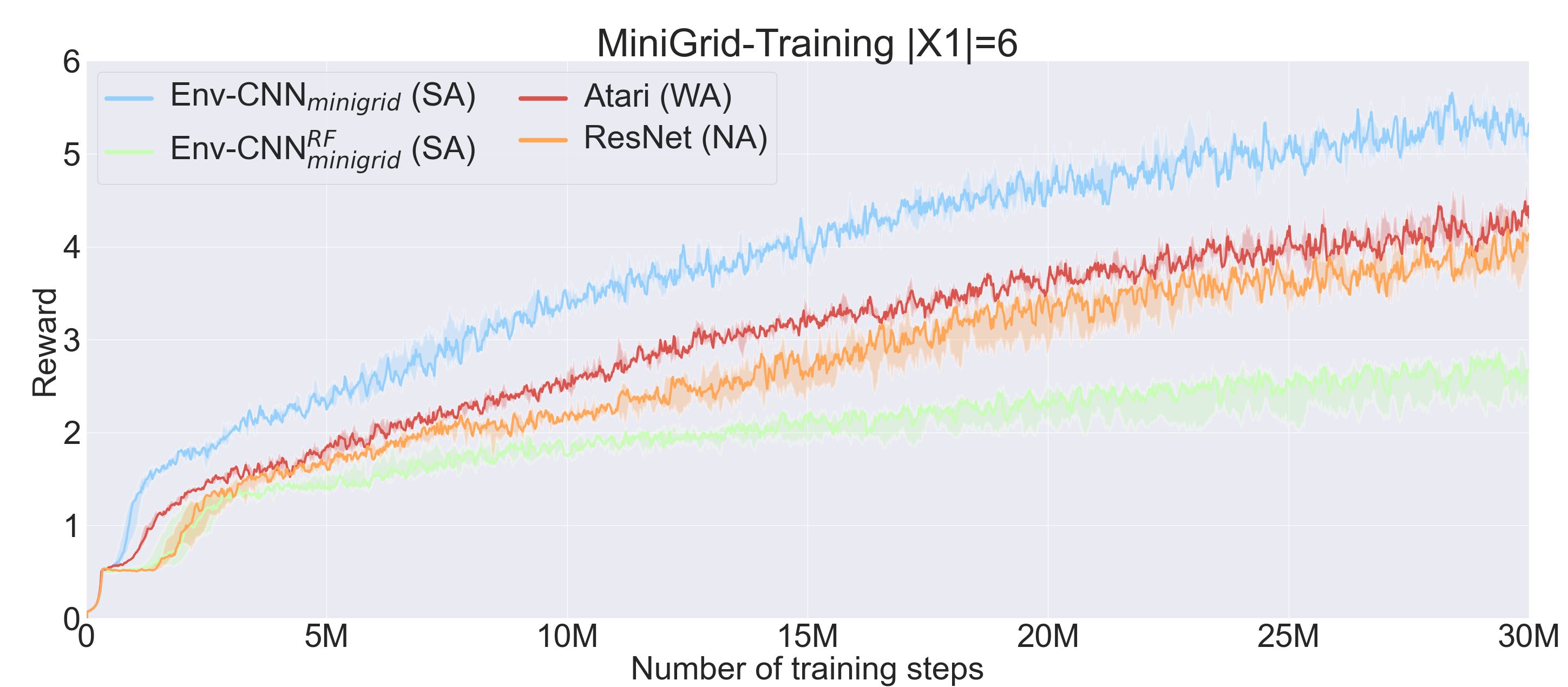}
    \centering \includegraphics[width=0.49\textwidth]{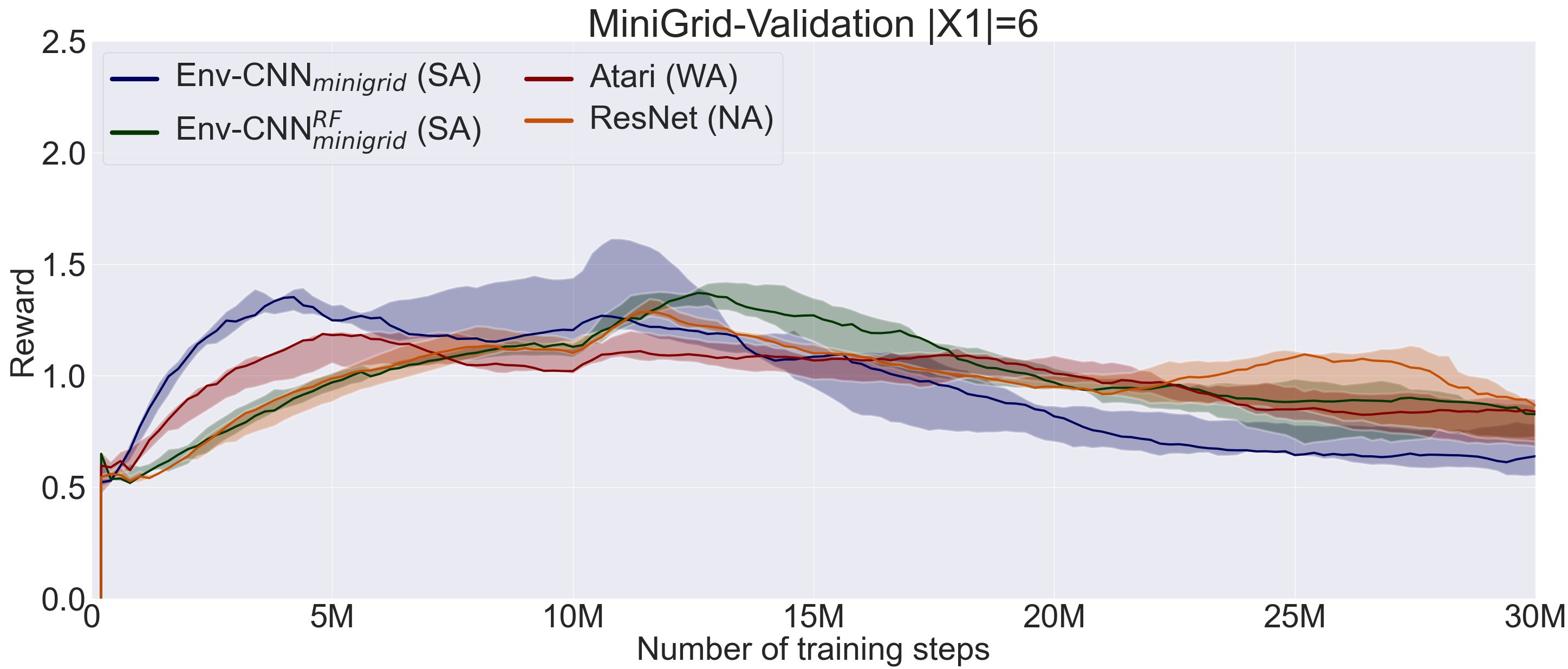}
    \vskip 0.15in
    \centering \includegraphics[width=0.49\textwidth]{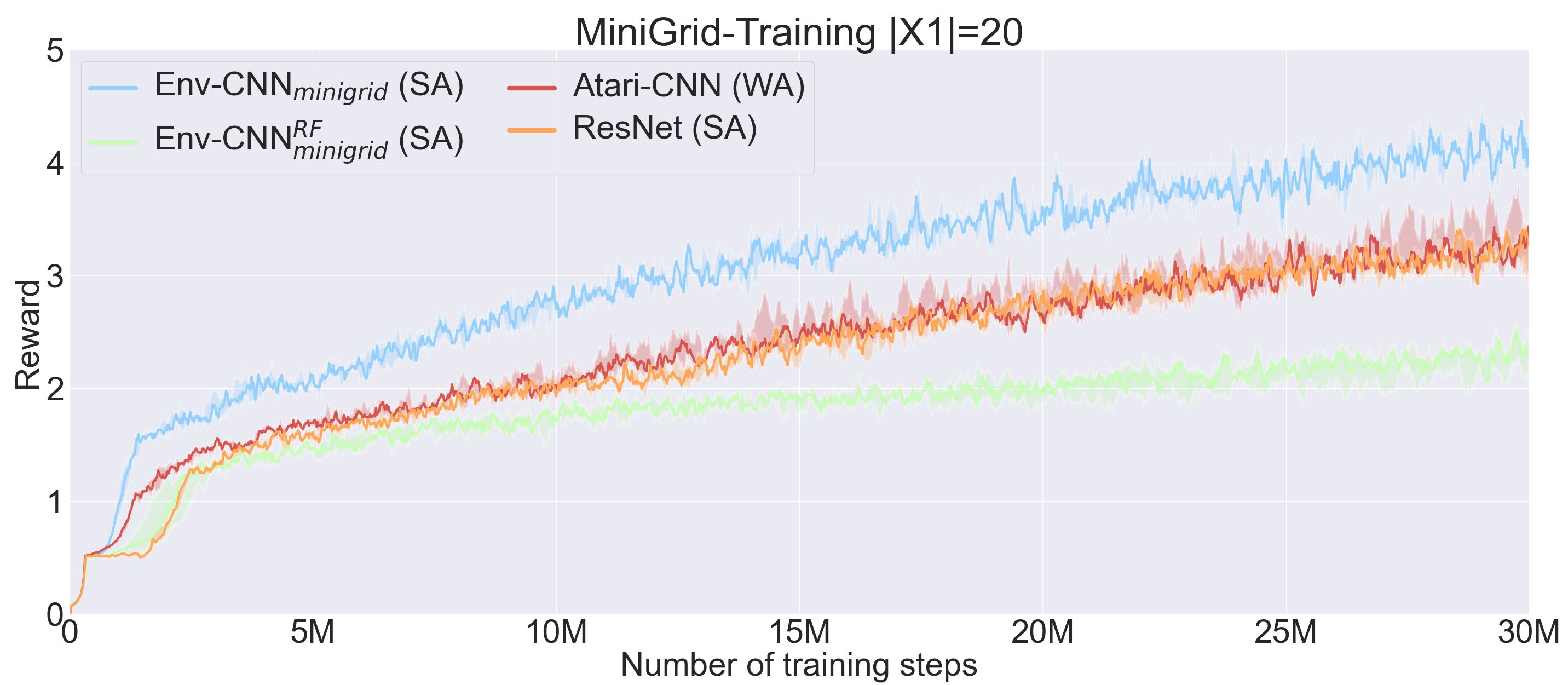}
    \centering \includegraphics[width=0.49\textwidth]{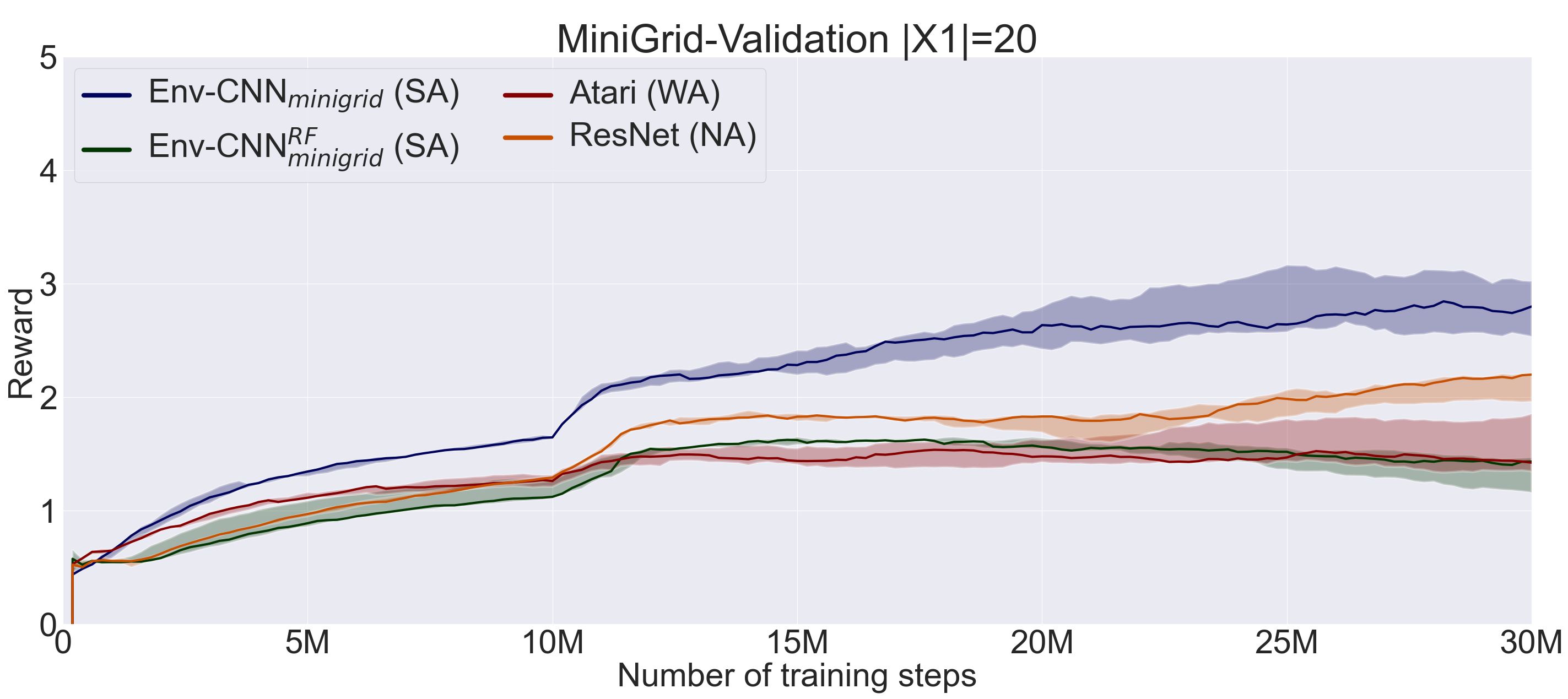}
    
    \caption{Train and validation plots with the different population sizes and 5 independent runs. Ligth colors refer to performance with training objects, whereas dark refer to validation objects. Lines refer to the median  while shaded areas represent the 25-75 quartile difference.}  \label{fig:training}
\end{figure*}

Figure~\ref{fig:training} shows the training and validation plots of the runs evaluated in Sec.~\ref{sec:experiments}. For each run we selected the weights that achieve the peak performance in the validation set.

\subsection{Hardware}

The various architectures were trained on computing clusters with GPUs such as Nvidia Tesla K80, Testla T4 or TITAN Xp. We employed Intel Xeon CPUs and consumed 7 GB of RAM per 5 independent runs working in parallel. Each experiment typically takes 2 days to train on this hardware. All the results presented in the main text and supplementary material are obtained from the aforementioned five independent runs per experiment.


\subsection{Complex instructions} \label{app: subsec: complex}
Table \ref{table:ComplexInst} contains the five complex instructions that we use to test our agent. The first two are the hardest specifications from \cite{andreas2017modular, toro2018teaching}. The three additional instructions were defined by aiming for combinations of negated tasks and inclusive/exclusive non-deterministic choices with non-atomic components together with sequential instances. We recall that operator $\cap$ can be defined in terms of sequential composition $;$ and non-deterministic choice $\cup$.
\input{utils/appendix_utils/table8}

%% file: utils/arquitectures.tex
\begin{figure*}[t] 
\centering
    \begin{adjustbox}{width=0.12\linewidth} 
     \begin{tikzpicture}
            [
             base/.style = {rectangle, rounded corners, draw=black,
                                       minimum width=1.3cm, minimum height=0.5cm,
                                       text centered, font=\sffamily},
            Map/.style = {rectangle, draw=black, fill=brown!10, minimum width=1.2cm, minimum height=1cm},
            Spec/.style = {rectangle, draw=black, fill=red!30, minimum width=1.2cm, minimum height=0.3cm},
            FullyC/.style = {base, fill=blue!15},
            Conv/.style = {base, fill=red!15},
            Encoder/.style = {base, fill=yellow!15},
            LSTM/.style = {base, fill=green!15},
            activation/.style = {base, minimum width=0.8 cm, minimum height=0.3cm, fill=orange!15,
                                       font=\ttfamily},
            output/.style = {base, minimum width=0.6cm, fill=white,
                                       font=\ttfamily},
            node distance=0.75cm,
            every node/.style={fill=white, font=\sffamily}, align=center
            ]
            
            \node (MapI) [Map]   {\scriptsize Obs};
            \node (SpecE) [Spec, node distance=0.7cm, above of=MapI]   {\scriptsize Task};
            \node (Enc) [Encoder, above of=SpecE, node distance=1cm]   {\scriptsize convolutional \\ \scriptsize layers };
            \node (Embed) [above of= Enc, node distance=1cm] {\scriptsize visual embedding};
            \node (Dense1)     [FullyC, above of=Embed]  {\scriptsize FC 128};
            \node (Relu1)     [activation, node distance=0.45cm, above of=Dense1]  {\scriptsize ReLu};
            \node (LSTM1)     [LSTM, above of=Relu1]  {\scriptsize LSTM 128};
            \node (Actor) [output, node distance=1cm, above right of=LSTM1] {\scriptsize $\pi(a_t)$};
            \node (Critic) [output, node distance=1cm, above left of=LSTM1] { \scriptsize $V_t$};
            
            
            \draw[->] (SpecE) -- (Enc);
            \draw[->] (Enc) -- (Embed);
            \draw[->] (Embed) -- (Dense1);
            \draw[->] (Relu1) -- (LSTM1);
            \draw[->] (LSTM1) -- (Actor);
            \draw[->] (LSTM1) -- (Critic);

             \node [below of= MapI, node distance=1.1cm]
            {
                \footnotesize Common layers \\
                \footnotesize Minecraft
            };
        \end{tikzpicture}
    \end{adjustbox}  
    \begin{adjustbox}{width=0.2\linewidth} 
     \begin{tikzpicture}
            [
             base/.style = {rectangle, rounded corners, draw=black,
                                       minimum width=1.3cm, minimum height=0.5cm,
                                       text centered, font=\sffamily},
            Map/.style = {rectangle, draw=black, fill=brown!10, minimum width=1.2cm, minimum height=1cm},
            Spec/.style = {rectangle, draw=black, fill=red!30, minimum width=1.2cm, minimum height=0.3cm},
            FullyC/.style = {base, fill=blue!15},
            Conv/.style = {base, fill=red!15},
            Encoder/.style = {base, fill=yellow!15},
            LSTM/.style = {base, fill=green!15},
            activation/.style = {base, minimum width=0.8 cm, minimum height=0.3cm, fill=orange!15,
                                       font=\ttfamily},
            output/.style = {base, minimum width=0.6cm, fill=white,
                                       font=\ttfamily},
            node distance=0.75cm,
            every node/.style={fill=white, font=\sffamily}, align=center
            ]
            
            \node (MapI) [Map]   {\scriptsize Obs};
            \node (Enc) [Encoder, above of=MapI, node distance=1.2cm]   {\scriptsize convolutional \\ \scriptsize layers};
            \node (EmbedV) [above of= Enc, node distance=1.2cm] {\scriptsize visual \\ \scriptsize embedding};
            
            \node (SpecE) [Spec, left of=MapI, node distance=1.8cm]   {\scriptsize Task};
            \node (LSTM1)     [LSTM, above of=SpecE, node distance=1.1cm]  {\scriptsize Bidirectional \\ \scriptsize LSTM 32};
            \node (EmbedT) [above of= LSTM1, node distance=1.2cm] {\scriptsize text \\ \scriptsize embedding};
            
            \node (SUM) [output, above right of=EmbedT, node distance=1.3cm]  { $+$};
             
            \node (Dense1)     [FullyC, above of=SUM]  {\scriptsize FC 128};
            \node (Relu1)     [activation, node distance=0.45cm, above of=Dense1]  {\scriptsize ReLu};
            \node (LSTM2)     [LSTM, above of=Relu1]  {\scriptsize LSTM 128};
            \node (Actor) [output, node distance=1cm, above right of=LSTM2] {\scriptsize $\pi(a_t)$};
            \node (Critic) [output, node distance=1cm, above left of=LSTM2] { \scriptsize $V_t$};
            
            
            \draw[->] (MapI) -- (Enc);
            \draw[->] (Enc) -- (EmbedV);
            \draw[->] (EmbedV) -- (SUM);
            \draw[->] (SpecE) -- (LSTM1);
            \draw[->] (LSTM1) -- (EmbedT);
            \draw[->] (EmbedT) -- (SUM);
            \draw[->] (SUM) -- (Dense1);
            \draw[->] (Relu1) -- (LSTM2);
            \draw[->] (LSTM2) -- (Actor);
            \draw[->] (LSTM2) -- (Critic);

             \node [below left of= MapI, node distance=1.4cm]
            {
                \scriptsize Common layers \\
                 \scriptsize Minigrid
            };
        \end{tikzpicture}
    \end{adjustbox}  
    \hfill
    \begin{adjustbox}{width=0.56\linewidth}
    \begin{tikzpicture}
            [
          base/.style = {rectangle, rounded corners, draw=black,
                                       minimum width=1.3cm, minimum height=0.5cm,
                                       text centered, font=\sffamily},
            Map/.style = {rectangle, draw=black, fill=brown!10, minimum width=1.2cm, minimum height=1cm},
            Spec/.style = {rectangle, draw=black, fill=red!30, minimum width=1.2cm, minimum height=0.3cm},
            FullyC/.style = {base, fill=blue!15},
            Conv/.style = {base, fill=red!15},
            Encoder/.style = {base, fill=yellow!15},
            LSTM/.style = {base, fill=green!15},
            activation/.style = {base, minimum width=0.8 cm, minimum height=0.3cm, fill=orange!15,
                                       font=\ttfamily},
            output/.style = {base, minimum width=0.6cm, fill=white,
                                       font=\ttfamily},
            node distance=0.75cm,
            every node/.style={fill=white, font=\sffamily}, align=center
            ]
            \node (MapI) [Map]   {\scriptsize Visual input};
            \node (ConvL1) [Conv, above of=MapI, node distance=1.4cm , minimum width=1.8cm]  {\scriptsize Conv. 3x3,\\\scriptsize stride 3, 16 ch.};
            \node (Relu1)     [activation, node distance=0.63cm, above of=ConvL1]  {\scriptsize ReLu};
            \node (ConvL2) [Conv, above of=Relu1, node distance=0.9cm]  {\scriptsize Conv. 3x3,\\ \scriptsize stride 3, 32 ch.};
            \node (Relu2)     [activation, node distance=0.63cm, above of=ConvL2]  {\scriptsize ReLu};
            \node (ConvL3) [Conv, above of=Relu2, node distance=0.9cm]  {\scriptsize Conv. 1x1,\\ \scriptsize stride 1, 1 ch.};

            \node (Output) [above of = ConvL3, node distance=0.9cm]
            {
                \scriptsize visual embedding
            };
            
            \draw[->] (MapI) -- (ConvL1);
            \draw[->] (Relu1) -- (ConvL2);
            \draw[->] (Relu2) -- (ConvL3);
            \draw[->] (ConvL3) -- (Output);
            
            \node (line1) [below of= MapI, node distance=1.2cm]
            {
                \footnotesize Env-CNN\textsubscript{minecraft}/\\
                \footnotesize Env-CNN\SPSB{RF}{minecraft}
            };
           
        \end{tikzpicture}

        \begin{tikzpicture}
            [
          base/.style = {rectangle, rounded corners, draw=black,
                                       minimum width=1.3cm, minimum height=0.5cm,
                                       text centered, font=\sffamily},
            Map/.style = {rectangle, draw=black, fill=brown!10, minimum width=1.2cm, minimum height=1cm},
            Spec/.style = {rectangle, draw=black, fill=red!30, minimum width=1.2cm, minimum height=0.3cm},
            FullyC/.style = {base, fill=blue!15},
            Conv/.style = {base, fill=red!15},
            Encoder/.style = {base, fill=yellow!15},
            LSTM/.style = {base, fill=green!15},
            activation/.style = {base, minimum width=0.8 cm, minimum height=0.3cm, fill=orange!15,
                                       font=\ttfamily},
            output/.style = {base, minimum width=0.6cm, fill=white,
                                       font=\ttfamily},
            node distance=0.75cm,
            every node/.style={fill=white, font=\sffamily}, align=center
            ]
            \node (MapI) [Map]   {\scriptsize Visual input};
            \node (ConvL1) [Conv, above of=MapI, node distance=1.4cm , minimum width=1.8cm]  {\scriptsize Conv. 2x2,\\\scriptsize stride 2, 16 ch.};
            \node (Relu1)     [activation, node distance=0.63cm, above of=ConvL1]  {\scriptsize ReLu};
            \node (ConvL2) [Conv, above of=Relu1, node distance=0.9cm]  {\scriptsize Conv. 2x2,\\ \scriptsize stride 2, 32 ch.};
            \node (Relu2)     [activation, node distance=0.63cm, above of=ConvL2]  {\scriptsize ReLu};
            \node (ConvL3) [Conv, above of=Relu2, node distance=0.9cm]  {\scriptsize Conv. 2x2,\\ \scriptsize stride 2, 32 ch.};
             \node (Relu3)     [activation, node distance=0.63cm, above of=ConvL3]  {\scriptsize ReLu};

            \node (Output) [above of = Relu3, node distance=0.7cm]
            {
                \scriptsize visual embedding
            };
            
            \draw[->] (MapI) -- (ConvL1);
            \draw[->] (Relu1) -- (ConvL2);
            \draw[->] (Relu2) -- (ConvL3);
            \draw[->] (Relu3) -- (Output);
            
            \node (line1) [below of= MapI, node distance=1.2cm]
            {
                \footnotesize Env-CNN\textsubscript{minigrid}/\\
                \footnotesize Env-CNN\SPSB{RF}{minigrid}
            };
           
        \end{tikzpicture}
        \begin{tikzpicture}
        [
        base/.style = {rectangle, rounded corners, draw=black,
                                       minimum width=1.3cm, minimum height=0.5cm,
                                       text centered, font=\sffamily},
            Map/.style = {rectangle, draw=black, fill=brown!10, minimum width=1.2cm, minimum height=1cm},
            Spec/.style = {rectangle, draw=black, fill=red!30, minimum width=1.2cm, minimum height=0.3cm},
            FullyC/.style = {base, fill=blue!15},
            Conv/.style = {base, fill=red!15},
            LSTM/.style = {base, fill=green!15},
            activation/.style = {base, minimum width=0.8 cm, minimum height=0.3cm, fill=orange!15,
                                       font=\ttfamily},
            output/.style = {base, minimum width=0.6cm, fill=white,
                                       font=\ttfamily},
            node distance=0.9cm,
            every node/.style={fill=white, font=\sffamily}, align=center
            ]
             \node (MapI) [Map]   {\scriptsize Visual input};
            \node (ConvL1) [Conv, above of=MapI, node distance=1.4cm , minimum width=1.8cm]  {\scriptsize Conv. 8x8,\\\scriptsize stride 4, 16 ch.};
            \node (Relu1)     [activation, node distance=0.63cm, above of=ConvL1]  {\scriptsize ReLu};
            \node (ConvL2) [Conv, above of=Relu1, node distance=0.9cm]  {\scriptsize Conv. 4x4,\\ \scriptsize stride 2, 32 ch.};
            \node (Relu2)     [activation, node distance=0.63cm, above of=ConvL2]  {\scriptsize ReLu};

            \node (Output) [above of = Relu2, node distance=0.7cm]
            {
                \scriptsize visual embedding
            };
            
            \draw[->] (MapI) -- (ConvL1);
            \draw[->] (Relu1) -- (ConvL2);
            \draw[->] (Relu2) -- (Output);

            \node [below of= MapI ,node distance=1.4cm]
            {
                \footnotesize Atari-CNN
            };
        \end{tikzpicture}
        \begin{tikzpicture}
        [
        base/.style = {rectangle, rounded corners, draw=black,
                                       minimum width=1.3cm, minimum height=0.5cm,
                                       text centered, font=\sffamily},
            Map/.style = {rectangle, draw=black, fill=brown!10, minimum width=1.2cm, minimum height=1cm},
            Spec/.style = {rectangle, draw=black, fill=red!30, minimum width=1.2cm, minimum height=0.3cm},
            FullyC/.style = {base, fill=blue!15},
            Conv/.style = {base, fill=red!15},
            LSTM/.style = {base, fill=green!15},
            MAX/.style = {base, fill=pink!15},
            Res/.style = {base, fill=green!30},
            activation/.style = {base, minimum width=0.8 cm, minimum height=0.3cm, fill=orange!15,
                                       font=\ttfamily},
            output/.style = {base, minimum width=0.6cm, fill=white,
                                       font=\ttfamily},
            node distance=0.9cm,
            every node/.style={fill=white, font=\sffamily}, align=center
            ]
            \node (MapI) [Map]   {\scriptsize Visual input};
            \node (ConvL1) [Conv, above of=MapI, node distance=1.4cm]  {\scriptsize Conv. 3x3, stride 1};
            \node (Max) [MAX, above of=ConvL1, node distance=0.9cm]  {\scriptsize Max 3x3, stride 2};
       
         \node (Res1) [Res, above of=Max]  {\scriptsize Residual Block};
          \node (Res2) [Res, above of=Res1]  {\scriptsize Residual Block};
          \node (Relu2) [activation, node distance=1cm, above of=Res2]  {\scriptsize ReLu};
          \node (Output) [above of = Relu2, node distance=0.7cm]{\scriptsize visual embedding};
        
        
        \draw[->] (MapI) -- (ConvL1);
        \draw[->] (ConvL1) -- (Max);
        \draw[->] (Max) -- (Res1);
        \draw[->] (Res1) -- (Res2);
        \draw[->,dashed] (Res2) -- (Relu2);
        \draw[->] (Relu2) -- (Output);
        \draw[thick,dashed] (0.5,1) -- (1.5,1);
        \draw[thick,dashed] (0.5,4.5) -- (1.5,4.5);
        \draw[thick,dashed] (1.5,1) -- (1.5,4.5);
         \node[node distance=1.4cm, below of= MapI]
            {
                \footnotesize   ResNet
            };
        \node [right of= Res1, node distance=2cm]
            {
                \scriptsize x3\\
                \scriptsize [64,64,32] ch.
            };
        \end{tikzpicture}
    \end{adjustbox}       %
        
\caption{The neural network architectures evaluated in Sec.~\ref{sec:experiments}. All the variants follow the structures depicted in the left for the corresponding settings. $\pi(a_t)$ and $V_t$ refer to the actor and critic layers respectively \cite{mnih2016asynchronous}. The different number of channels in the last layer of the Env-CNN architectures is done to provide evidence that such feature does not affect how the agent generalises.}
\label{fig:NNarchitectures}
\end{figure*}

%% file: utils/appendix_utils/table8.tex
\begin{table*}[h]
  \centering
  \small
  \begin{tabular}{M{5cm}M{8cm}}
\toprule
  TTL  & Intuitive meaning\\
\midrule
$((iron ; workbench) \cap wood) ; toolshed ; axe $ & Get iron and then use workbench, also get wood. Then use the toolshed. Later use the axe.\\
\midrule        
$(wood \cap iron); workbench$     &     Get wood and iron. Then use the workbench\\
 \midrule               
($grass\sim~) ; grass ; (workbench \cup toolshed)$ &  Get an object different from grass. Then get grass. Later use either the workbench or the toolshed. \\
 \midrule      
$((workbench\sim) \cap (toolshed\sim~)); toolshed$     &  Use an object different from the workbench and use another object different from the toolshed. Then use the toolshed\\
 \midrule      
$((wood ; grass) \cup (iron; axe)); workbench; (toolshed \sim~)$     &  Get either wood and then grass or iron and an axe. Then use the workbench. Later use something different from the toolshed.\\

 \bottomrule
\end{tabular}
\caption{Complex instructions used in Table~\ref{table:ComplexT} in the main text. The 6 objects included in these instructions belong to the test set ($\mathcal{X}_3$) in the Minecraft-inspired environment.}\label{table:ComplexInst}

\end{table*}

%% file: Appendix/SModule.tex
In this section we include the pseudo-code of the of the symbolic module, whose procedures are explained in Sec.~\ref{subsec:Smodule} of the main text.

Algorithm~\ref{alg:SymbM} in the main text details the Symbolic Module (SM), which given a TTL instruction $\phi$, decomposes the instruction into atomic tasks $T$ to be solved by the Neural Module (NM). Algorithm~\ref{alg:extractor} details the extractor $\mathcal{E}$, an internal function of the SM that given the instruction formula generates a set $\mathcal{K}$ of sequences of tasks that satisfy $\phi$. Algorithm~\ref{alg:prog} refers to the progression function $\mathcal{P}$, which updates $\mathcal{K}$ according to the true evaluation $p$ given by the internal labelling function $\mathcal{L}_I$ that indicates to the SM how the previous task was solved.

  \input{utils/algorithms/Extractor}
  
  \input{utils/algorithms/Progression}

%% file: utils/algorithms/Extractor.tex
\begin{algorithm}[t]
\caption{Extractor function that obtains the list-of-lists with the possible sequences of tasks.}
\label{alg:extractor}
\begin{algorithmic}[1]
\STATE {\bfseries Function $\mathcal{E}$, Input:}  $\phi$
\STATE Initialize the list $\mathcal{K}$ with an empty sequences of tasks $Seq_0$
     \FOR {each atomic task $T \in \phi$}
        \IF { $ T$ is not a non-deterministic choice}
            \FORALL{$Seq \in \mathcal{K}$}
                \STATE $Seq$.append$( T)$
                \ENDFOR 
        \ELSE 
            \STATE Save the first element:  $FC \leftarrow  T[0]$
            \STATE Save the last element: $SC \leftarrow  T[-1]$
            \STATE Nseqs $\leftarrow length(\mathcal{K})$
             \FORALL{$Seq \in \mathcal{K}$}
                \STATE Generate a clone:  $Seq' \leftarrow  Seq$
                \STATE $\mathcal{K}$.append($Seq'$)
            \ENDFOR
            \FOR{$i$ in \textbf{range}$(length(\mathcal{K}))$}
                \IF{$i <$ Nseqs}
                    \STATE $\mathcal{K}[i]$.append$(FC)$
                \ELSE
                    \STATE $\mathcal{K}[i]$.append$(SC)$
                \ENDIF
            \ENDFOR
        \ENDIF    

    \ENDFOR
     \RETURN $\mathcal{K}$
\end{algorithmic}
\end{algorithm}



%% file: utils/algorithms/Progression.tex
\begin{algorithm}[t]
\caption{Progression function that returns the next task to be solved.}
\label{alg:prog}
\begin{algorithmic}[1]
\STATE {\bfseries Function {$\mathcal{P}$}, Input:}{$\mathcal{K}, p$}
    \IF{$ p \neq \emptyset$}
        \FORALL{$Seq \in \mathcal{K}$}
            \IF{$p$ fulfills $Seq[0]$}
                \STATE $Seq$.pop($Seq[0]$)
            \ELSE
                \STATE $\mathcal{K}$.pop($Seq$)
            \ENDIF
        \ENDFOR
    \ENDIF
    
    \STATE $ T \leftarrow \emptyset$
   
     \IF{$\mathcal{K} \text{ is not } \emptyset$}
        \STATE Select the head $T$ of a sequence in $\mathcal{K}$:
        \IF{$T \equiv \alpha$}
            \STATE //Aiming to form a non-deterministic choice
            \FORALL{$Seq \in \mathcal{K}$} 
                \IF{$T==Seq[0]$}
                    \STATE \textbf{continue}
                \ELSIF{$Seq[0] \equiv \alpha$}
                        \STATE $ T \leftarrow T \lor Seq[0]$
                        \STATE \textbf{return} $ T$
                \ENDIF
            \ENDFOR
        \ENDIF
    \ENDIF
    \RETURN $T$
\end{algorithmic}
\end{algorithm}

%% file: Appendix/control.tex
This section includes the results with disjunction anticipated in Sec.~\ref{sec:experiments}. 
\input{utils/appendix_utils/table6}

Table~\ref{table:DisjBCMs} shows the results of the different architectures when following reliable, i.e., pointing to the right object, and deceptive, i.e., pointing to the wrong object, zero-shot instructions with a non-deterministic choice. Instructions labelled
``Disjunction 1\textsuperscript{st} choice" refer to those where only the first object within the instruction is present in the map. ``Disjunction 2\textsuperscript{nd} choice" is a similar setting with the second object. In ``Disjunction 2 choices" both objects are present in the map. As detailed in the main text, agents learning compositionally should be able to follow the zero-shot formulae and consequently do well with reliable instructions while poorly with deceptive ones.

%% file: utils/appendix_utils/table6.tex
\begin{table*}[t]
\renewcommand{\arraystretch}{1}
  \centering
    \mycfs{8.2}
   \begin{tabular}{m{1.6cm}P{0.8cm}cccccccc}
    \toprule
\textit{Minecraft} &  $|\mathcal{X}_{1}|$ &\multicolumn{2}{c}{Atari-CNN (\textbf{NA})} &  \multicolumn{2}{c}{ResNet (\textbf{WA})}  & \multicolumn{2}{c}{Env-CNN\textsubscript{minecraft} (\textbf{SA})} 
  & \multicolumn{2}{c}{Env-CNN\SPSB{RF}{minecraft} (\textbf{SA})}\\
\cmidrule(r){3-4}
\cmidrule(r){5-6}
\cmidrule(r){7-8}
\cmidrule(r){9-10}
Instruction  & & Reliable & Deceptive & Reliable & Deceptive & Reliable & Deceptive &  Reliable & Deceptive  \\
\cmidrule(r){1-2}
Disjunction & 6  & 3.39 $(0.10)$ & 2.12$(0.02)$
        &	2.89$(0.30)$ & 2.04$(0.09)$ & \textbf{3.01}$(1.37)$ & \textbf{0.76$(0.55)$} & 
        \textbf{2.58}$(1.81)$ &\textbf{1.24$(0.51)$}\\

 1\textsuperscript{st} choice & 20 & 3.60$(0.0)$ & 2.18$(0.03)$&     
    2.67$(1.20)$ & 1.77$(1.03)$	& \textbf{5.10}$(1.17)$ &	\textbf{0.47}$(0.46)$  &  
    2.93$(1.64)$ & 1.02$(0.82)$\\

\cmidrule(r){1-2}
Disjunction & 6  & 3.25 $(0.23)$ & 2.13$(0.03)$
        &	2.83$(0.48)$ & 2.02$(0.17)$ &\textbf{ 3.05$(1.35)$} & \textbf{0.57$(0.46)$} & 
        \textbf{2.63}$(1.67)$ &\textbf{1.05$(0.51)$}\\

 2\textsuperscript{nd} choice & 20 & 3.60$(0.09)$ & 2.17$(0.02)$&     
    2.89$(1.28)$ & 1.65$(0.20)$	& \textbf{4.91}$(1.53)$ &	\textbf{0.53}$(0.61)$  &  
    2.96$(1.91)$ & 1.04$(0.72)$\\
\cmidrule(r){1-2}    
Disjunction & 6  & 2.56 $(0.23)$ & 2.59$(0.57)$
        &	2.29$(0.12)$ & 2.51$(0.69)$ & \textbf{2.23$(4.79)$} & \textbf{0.57$(0.29)$} & 
        2.54$(0.67)$ &1.37$(0.06)$\\

 2 choices & 20 & 2.63 $(0.23)$ & 2.68$(0.57)$
        &	2.16$(0.12)$ & 2.23$(0.69)$ & \textbf{3.53$(4.79)$} & \textbf{0.44$(0.29)$} & 
        \textbf{2.82$(0.67)$ }& \textbf{1.07$(0.90)$}\\\\
    
\midrule    

\textit{Minigrid} &  $|\mathcal{X}_{1}|$ &\multicolumn{2}{c}{Atari-CNN (\textbf{WA})} &  \multicolumn{2}{c}{ResNet (\textbf{NA})}  & \multicolumn{2}{c}{Env-CNN\textsubscript{minigrid} (\textbf{SA})} 
  & \multicolumn{2}{c}{Env-CNN\SPSB{RF}{minigrid} (\textbf{SA})}\\
\cmidrule(r){3-4}
\cmidrule(r){5-6}
\cmidrule(r){7-8}
\cmidrule(r){9-10}
Instruction   & & Reliable & Deceptive & Reliable & Deceptive & Reliable & Deceptive &  Reliable & Deceptive  \\
\cmidrule(r){1-2}
Disjunction & 6  & \textbf{2.48 $(0.57)$} & \textbf{0.93}$(0.30)$
        &	3.73$(0.61)$ & 3.64$(1.29)$ &\textbf{ 2.09$(1.20)$} & \textbf{0.95$(0.64)$} & 
        \textbf{.36}$(0.88)$ &\textbf{0.79$(0.48)$}\\

 1\textsuperscript{st} choice & 20 &\textbf{ 2.99$(0.86)$} & \textbf{1.01$(0.36)$}&     
    \textbf{6.14$(0.82)$} &\textbf{ 1.49$(0.83)$}	& \textbf{2.61}$(0.58)$ &	\textbf{0.39}$(0.06)$  &  
    2.06$(0.77)$ & 1.31$(0.34)$\\

\cmidrule(r){1-2}    
Disjunction & 6  & \textbf{2.83 $(0.93)$} & \textbf{0.90}$(0.41)$
        &	3.52$(0.79)$ & 3.86$(1.30)$ &\textbf{ 1.68$(1.25)$} & \textbf{0.75$(0.27)$} & 
        \textbf{2.81}$(0.79)$ &\textbf{0.88$(0.55)$}\\

 2\textsuperscript{nd} choice & 20 & 2.11$(1.42)$ & 1.06$(0.32)$&     
    2.60$(0.74)$ & 2.37$(1.03)$	& \textbf{1.81}$(0.52)$ &	\textbf{0.50}$(0.08)$  &  
    1.91$(0.76)$ & 1.28$(0.4)$\\
    
\cmidrule(r){1-2}
Disjunction & 6  & \textbf{4.21} $(0.46)$ & \textbf{0.84} $(0.50)$
        &	4.02$(0.39)$ & 3.68$(0.59)$ & \textbf{2.67} $(1.49)$ & \textbf{ 0.79}$(0.54)$ & 
        \textbf{4.06}$(0.74)$ &\textbf{0.63$(0.44)$}\\

2 choices  & 20 & \textbf{3.42}$(0.97)$ & \textbf{1.14}$(0.48)$&     
    \textbf{5.03$(0.59)$} & \textbf{ 1.75$(0.77)$}	& \textbf{3.13}$(0.47)$ &	\textbf{0.37}$(0.09)$  &  
    2.72$(0.97)$ &1.58$(0.38)$\\
    
 \bottomrule
\end{tabular}
\normalsize
\caption{Complete results in BCMs with disjunctions. Results are bolded where performance with reliable instructions is greater or equal than two times the respective performance with deceptive instructions. }\label{table:DisjBCMs}

\end{table*}

%% file: Appendix/StruggNegation.tex
This section section aims to bring light to an apparent contradiction between our empirical results and those from the state-of-the-art about generalisation with negation \cite{hill2020environmental}. 

In Sec.~\ref{sec:experiments} in the main text, we observe that the architecture of the CNN plays a key role in the ability of agents to generalise when the number of instructions is limited. Still, in Table~\ref{table:BCMs} we see that the ResNet shows good generalisation in both settings when learning from 20 negated instructions. This seems to contradict the results from \citet{hill2020environmental} where the same ResNet struggles to generalise with 40 negated instructions. Nevertheless, we find that this is caused by the particular form that neural networks overfit to the negation operator. 

In particular, consider the TTL task {\em copper $\sim$}, which intuitively means ``get something different from copper". When agents overfit they no longer interpret the negation operator as a separated element, but rather as if the operator was part of the object's name. Consequently, for an overfitting agent, {\em copper $\sim$} is a single word that represents all the objects of the training set except copper. This overfitting phenomena produces policies that, when facing non-training objects, are no longer able to interpret negation correctly, leading to a behavior different from what an agent learning systematically is expected to do. We find that this phenomena, first described in \citet{hill2020environmental}, happens only {\em after} agents overfit to the training set. This means that through a validation set, we can perform early stopping -- as we do in Sec.~\ref{sec:experiments} -- and use policies that correctly generalise to new instructions without requiring a larger number of training instructions. 

Table~\ref{table:Overfitted} presents the results of an ablation study about the use of a validation set with a ResNet agent trained until convergence in the Minecraft large training set. We see that there is little difference with positive instructions, but results are completely opposite in the case of negated tasks, that go from an agent correctly generalising negation -- when applying a validation set and early stopping -- to one that does the opposite from what the specification requires -- when no validation set is used --. Consequently, {\em the struggle with negated instructions in previous research seems to be motivated by the different ways deep learning overfits when learning different abstract operators}.

\begin{table*}[t]

  \centering
  \small
\begin{tabular}{lcccc} 
\toprule
 ResNet ($ |\mathcal{X}_1|=20$) & Reliable - No validation & Deceptive - No validation & Reliable - Validation & Deceptive - Validation\\
\midrule
Positive instructions & \textbf{3.08}(1.02) & \textbf{1.34}(0.65) & \textbf{3.30}(0.49) & \textbf{1.10}(0.81)\\
Negated instructions & 1.18(0.46) & 1.61(0.72) & \textbf{3.09}(0.49) & \textbf{1.26}(0.93)\\
\bottomrule
\end{tabular}
\caption{Results from a ResNet in the Minecraft setting when using a validation set or not, i.e., the latter are the results from an agent that is overfitting to the training set. We see that the different behaviours with positive and negative when the agent is overfitting align with the results from \citet{hill2020environmental}. Results are bolded where performance with reliable instructions is greater or equal than two times the respective performance with deceptive instructions. }\label{table:Overfitted}
\end{table*}

%% file: Appendix/operator.tex
Below we include additional details about the requirements we faced
when aiming a neuro-symbolic agent that generalises the negation and non-deterministic choice operators.

\subsection{Negation} \label{subsec:Appendixnegation}
When learning positive and negative instances concurrently using visual instructions, we noticed that the position of the negation operator impacted on the agent's behavior in training. Specifically, placing the negation symbol after the negated object helps the agent to learn both types of instances concurrently, while, when placed otherwise, the agent focus only on the positive tasks first, which can lead to saddle points. 

We believe that this result is not related to linguistic processing. Note that all of our architectures in Minecraft process the specification as a whole, as part of the observational input processed by the convolutional layer. Thus, the order is not important from the point of view of temporal processing. We believe the difference was caused because affirmative formulas are represented with one symbol, e.g., "get wood" is represented as "wood"; while negation is represented by two symbols, e.g., "get an object different from wood" as "wood $\sim$". By placing the negation operator after the object we force the neural network to "observe" the first two inputs to differentiate affirmative tasks from negative ones. An alternative approach that could have a similar effect is adding an explicitly positive operator so that both negative and positive formulas have the same length.

\subsection{Non-deterministic Choice}


Training the agent to effectively learn non-deterministic choice $\cup$ required episodes where only one of the two disjuncts of the non-deterministic choice was present, as well as some other episodes with the two objects present simultaneously. Intuitively, if we have the instruction "get axe or get gold", there should be training episodes where gold is present in the map, but there are no axes, and similarly, other maps where there are axes, but there is not gold. This prevents the agent from biasing with the object that appears either in the first or second position of the instruction. Additionally, the agent should also interact with episodes where both objects are present. Otherwise, the policy network of the agent will allocate a similarly high probability to go after the two objects, instead of to the one that is closer or "safer", i.e, far from non-desired objects.